\documentclass[runningheads]{llncs}
\usepackage[T1]{fontenc}
\usepackage{graphicx}
\usepackage{amsmath,amsfonts,amssymb,mathtools}

\usepackage{booktabs}

\usepackage{subcaption}
\usepackage{enumitem}

\usepackage[numbers,sort&compress]{natbib}
\usepackage{amsmath}
\usepackage{bbm}
\usepackage[ruled]{algorithm2e}

\graphicspath{{./images/}}

\usepackage[dvipsnames]{xcolor}
\usepackage{tikz}

\usepackage[disable]{todonotes}
\usepackage[pdftex,pdfpagelabels,bookmarks,hyperindex,hyperfigures,bookmarksnumbered]{hyperref}

\usepackage[charter]{mathdesign}
\usepackage{tikz}
\usetikzlibrary{decorations.pathreplacing,calligraphy}
\usetikzlibrary{tikzmark}
\usetikzlibrary{arrows.meta}

\begin{document}

\newcommand{\ourmethod}{revised ct-SNE\xspace}
\newcommand{\Ourmethod}{Revised ct-SNE\xspace}

\newcommand{\pstart}[1]{\noindent\textbf{#1.}}

\newcommand{\kl}[2]{\ensuremath{\text{KL}(#1\,\|\,#2)}}
\newcommand{\X}{\ensuremath{\mathbf{X}}}
\newcommand{\R}{\ensuremath{\mathbf{R}}}
\newcommand{\Q}{\ensuremath{\mathbf{Q}}}
\newcommand{\Y}{\ensuremath{\mathbf{Y}}}
	
\definecolor{blue}{RGB}{31, 119, 180}
\definecolor{orange}{RGB}{255, 127, 14}
\definecolor{smartseq2}{HTML}{B80058}
\definecolor{indrop}{HTML}{EBAC23}
\definecolor{celseq}{HTML}{006E00}
\definecolor{fluidigmc1}{HTML}{00BBAD}
\definecolor{ductal}{RGB}{0, 140, 249}
\definecolor{acinar}{RGB}{0, 110, 0}
\definecolor{alpha}{RGB}{235, 172, 35}
\definecolor{beta}{RGB}{184, 0, 88}

\newcommand{\tikzcircle}[1]{\tikz[baseline=-0.8ex]\draw[#1, fill=#1, opacity=0.7, radius=2.8pt] (0,0) circle ;}%
\newcommand{\emptycirc}{\tikz[baseline=-0.8ex] \draw[line width=0.6pt] circle(3pt);}
\newcommand{\tikzsquare}[1]{\tikz\filldraw[#1, fill=#1, opacity=0.7] (0,0) rectangle ({5.6pt},{5.6pt});}%
\newcommand{\expnumber}[2]{{#1}\mathrm{e}{#2}}
\newcommand{\rnx}{\ensuremath{R_{NX}}\xspace}
\title{Revised Conditional t-SNE:\\Looking Beyond the Nearest Neighbors}

\author{Edith Heiter\inst{1} \and Bo Kang\inst{1} \and Ruth Seurinck\inst{12} \and Jefrey Lijffijt\inst{1}}%

\authorrunning{E. Heiter et al.}
\institute{Ghent University, Belgium\\
\email{\{edith.heiter,bo.kang,ruth.seurinck,jefrey.lijffijt\}@ugent.be} \and
VIB Center for Inflammation Research, Belgium}
\maketitle              %
\begin{abstract}
Conditional t-SNE (ct-SNE) is a recent extension to t-SNE that allows removal of known cluster information from the embedding, to obtain a visualization revealing structure beyond label information. This is useful, for example, when one wants to factor out unwanted differences between a set of classes.
We show that ct-SNE fails in many realistic settings, namely if the data is well clustered over the labels in the original high-dimensional space.
We introduce a revised method by conditioning the high-dimensional similarities instead of the low-dimensional similarities and storing within- and across-label nearest neighbors separately. This also enables the use of recently proposed speedups for t-SNE, improving the scalability.
From experiments on synthetic data, we find that our proposed method resolves the considered problems and improves the embedding quality. On real data containing batch effects, the expected improvement is not always there. We argue revised ct-SNE is preferable overall, given its improved scalability. The results also highlight new open questions, such as how to handle distance variations between clusters.

\end{abstract}
\section{Introduction}

\begin{figure}[t]
	\centering
	\begin{subfigure}{.33\textwidth}
		\centering
		\includegraphics[height=31mm]{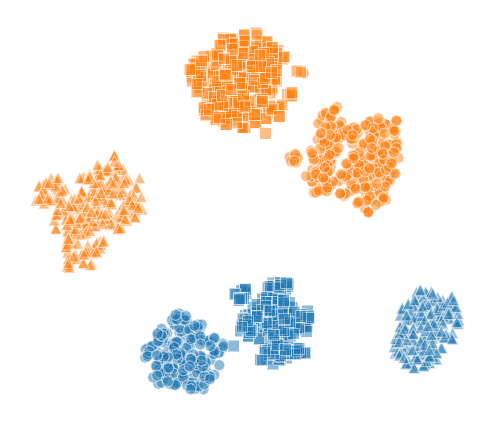}
		\caption{t-SNE\label{fig:tsne_intro}}
	\end{subfigure}\hfill
	\begin{subfigure}{.33\textwidth}
		\centering
		\includegraphics[height=31mm]{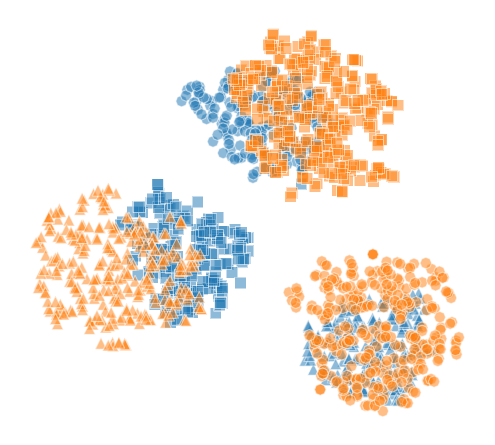}
		\caption{Conditional t-SNE\label{fig:ctsne_intro}}
	\end{subfigure}\hfill
	\begin{subfigure}{.33\textwidth}
		\centering
		\includegraphics[height=31mm]{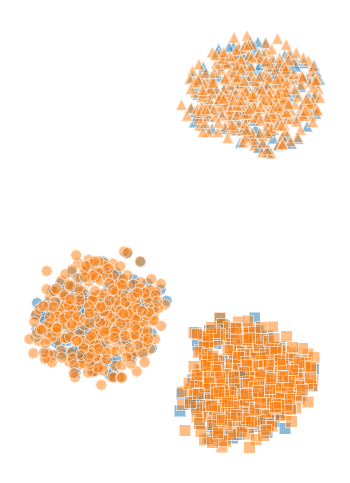}
		\caption{\Ourmethod\label{fig:fastctsne_intro}}
	\end{subfigure}
	\caption{Illustration of label information used by ct-SNE and \ourmethod. (a) t-SNE shows the labeled data consists of several clusters. Provided with the class labels \textbf{\textcolor{blue}{blue}} and \textbf{\textcolor{orange}{orange}}, \emph{ct-SNE (b) merges points from both colors--but the wrong shapes}, while \ourmethod (c) shows the three expected clusters.\label{fig:embeddings_intro}}
\end{figure}

\pstart{Motivation} t-distributed Stochastic Neighbor Embedding (t-SNE) \cite{van2008visualizing} is widely used to compute low-dimensional visualizations for high-dimensional data. Conditional t-SNE (ct-SNE) \cite{kang2021conditional} is an extension of t-SNE that allows to factor out prior knowledge from the embedding. Providing discrete labels for all data points to ct-SNE allows same-labeled points to be embedded further apart than their distances would require---ideally revealing complementary structure present in the data. 

We illustrate the idea of conditional t-SNE using a synthetic dataset (n=1500, d=10)  in Figure \ref{fig:embeddings_intro}a. The data is generated such that each point belongs to one of two clusters in dim 1-4 (\textbf{\textcolor{blue}{blue}},  \textbf{\textcolor{orange}{orange}}) and one of three clusters in dim 5-6 ($\emptycirc, \triangle, \square$). The remaining four dimensions are Gaussian noise. The t-SNE embedding in \ref{fig:embeddings_intro}a shows a separate cluster for each class label combination. If we already know about the clustering in dim 1-4, we could provide the labels \textbf{\textcolor{blue}{blue}}/\textbf{\textcolor{orange}{orange}} to ct-SNE. Ideally, this would reveal the remaining structure in the data: Three clusters in dimension 5-6 ($\emptycirc, \triangle, \square$). Figure \ref{fig:embeddings_intro}b shows that ct-SNE wrongly merges the clusters, i.e., \tikzcircle{orange} are merged with \textcolor{blue}{$\blacktriangle$}, while \ourmethod does show the correct clusters (Figure \ref{fig:embeddings_intro}c).

\vspace{4pt}\pstart{Use cases of ct-SNE} (Revised) ct-SNE retains the unsupervised nature of t-SNE while adding supervision through labels to explicate what is \emph{not} the target. This stands in contrast to supervised dimensionality reduction methods that incorporate label information to improve downstream prediction tasks, for example by increasing class separation in low-dimensional embeddings (see, e.g., \cite{vu2021hct, Bodt2019ClassawareTC}). %
As such, revised ct-SNE is useful in a situation where t-SNE is useful and when additionally there is known unwanted structure in the data. This may be in an iterative EDA setting, when clusters are identified, explored, and labeled, after which the user wants to explore further. Another setting is when label information about prominent structure in the data is available a priori, and this information acts as a confounder \citep{kang2021conditional}.

The presence of undesired and known class separation occurs for example with biological data containing single-cell RNA samples from various sources. The unwanted class separation is called the \textit{batch effect} and can be seen as variation in the data that does not have a biological explanation. It often occurs when combining samples from different organisms, tissues, or when cells have been sequenced with different technologies.
For this setting also other t-SNE variants have been proposed. \citet{Policar2021} suggest to embed one dataset (batch) using t-SNE and use this as a reference embedding. The other datasets are then embedded sample by sample on top of the initial embedding. This is different from our (revised) ct-SNE, in that it prevents any interaction of same-labeled samples by design. In addition, (revised) ct-SNE allows the user to tune the degree of class separation.

\vspace{4pt}\pstart{Contributions} In this paper we provide a thorough analysis of the root cause of ct-SNE's failures. We identify that the approximation of high-dimensional similarities discards essential structural information. In addition, the asymmetry of the KL-divergence hinders ct-SNE in achieving its goal. To overcome these limitations, we propose two modifications. First, we compute distances for same-labeled and differently-labeled neighbors separately. Second, we condition the high-dimensional instead of the low-dimensional similarities. Finally, we implemented \ourmethod into FIt-SNE \cite{linderman2019fast} which leads to a considerable speed-up.

\section{Background: Conditional t-SNE\label{sec:ctsne}}
In this section we review ct-SNE and point out details that might negatively affect the embedding quality.
The objective of ct-SNE is to embed a dataset $X\in \mathbb{R}^{n \times d}$ to a lower dimension $Y\in \mathbb{R}^{n \times d'}$ with $d' \ll d$ by minimizing the Kullback-Leibler (KL) divergence between pairwise similarities in the high (HD) and low-dimensional (LD) space. The HD similarities
\begin{align}
	p_{j|i} &=
		\frac{\exp(-\|x_i - x_j\|^2/2\sigma_i^2)}{\sum_{k \neq i} \exp(-\|x_i - x_k\|^2/2\sigma_i^2)}\,,\qquad p_{ij} = \frac{p_{i|j} + p_{j|i}}{2n}\nonumber
\end{align}
are defined with a point-specific kernel bandwidth $\sigma_i$ that depends on the density of the neighborhood around each point. It is computed by binary search such that each similarity distribution $p_i$ has the same user-defined perplexity \textit{u}.
The LD similarities are based on a t-distribution
\begin{align*}
	q_{ij} = \frac{\left(1+\|y_i - y_j\|^2\right)^{-1}}{\sum_{k \neq l} \left(1 + \|y_k - y_l\|^2\right)^{-1}}\,.
\end{align*}

In ct-SNE, the LD similarities are conditioned on the label matrix $\Delta \in \{0,1\}^{n\times n}$, based on the idea that $q_{ij|\Delta}$ should be higher for pairs of points with the same label ($\delta_{ij} = 1$) than for points with a different label ($\delta_{ij} = 0$). The conditional LD similarities are defined as
\begin{align*}
	r_{ij} = q_{ij|\Delta} = 
	\begin{cases}
		\alpha q_{ij}/U \qquad \text{if } \delta_{ij} = 1\\
		\beta q_{ij}/U \qquad \text{if } \delta_{ij} = 0
	\end{cases}
\end{align*}
and normalized with $U = \alpha \sum_{k\neq l:\delta_{kl} = 1} q_{kl} + \beta \sum_{k\neq l: \delta_{kl} = 0} q_{kl}$. We refer to the original publication for the detailed derivation and the exact relationship between the parameters $\alpha > 1 > \beta > 0$. Minimizing $\kl{p}{r}$ with $\alpha > \beta$ requires differently-labeled points to be embedded closer to each other to still match their pairwise HD similarity.

We illustrate the effect of ct-SNE on its gradient
\[
\nabla_{y_i}\kl{p}{r} = 4\sum_{j\neq i} \left(\tikzmarknode{attr}{p_{ij}q_{ij}Z}(y_i - y_j)- {\tikzmarknode{rep}{\frac{\delta_{ij}\alpha + (1-\delta_{ij})\beta)}{U} q_{ij}^2Z}}(y_i - y_j)\right)
\begin{tikzpicture}[overlay,remember picture,gray,>=Stealth]
	\draw [decorate, decoration = {brace,mirror}] (attr.south west) --  (attr.south east);
	\draw [decorate, decoration = {brace,mirror}] (rep.south west) --  (rep.south east);
	\node[below right=.05cm and -1cm of attr] {\small{attractive force}};
	\node[below=.05cm of rep] {\small{repulsive force}};
\end{tikzpicture}
\vspace{.2cm} 
\]
with $Z = \sum_{k\neq l}(1+\|y_k - y_l\|^2)^{-1}$. The attractive part will pull neighboring points $i$ and $j$ closer together while the repulsive part pushes all points apart. 
First, we note that ct-SNE increases (decreases) the repulsive force between same-labeled (differently-labeled) points, while the attractive forces are the same as in t-SNE. To speed up the computation of the attractive forces \citet{van2014accelerating} proposed to exploit the fast decay of the Gaussian kernel and retain only the similarities $p_{\cdot|i}$ for the set of  $|\mathcal{N}_i| = 3u$ nearest neighbors, where $u$ is the perplexity.

\begin{figure}[t]
	\centering
	\includegraphics[width=.7\textwidth]{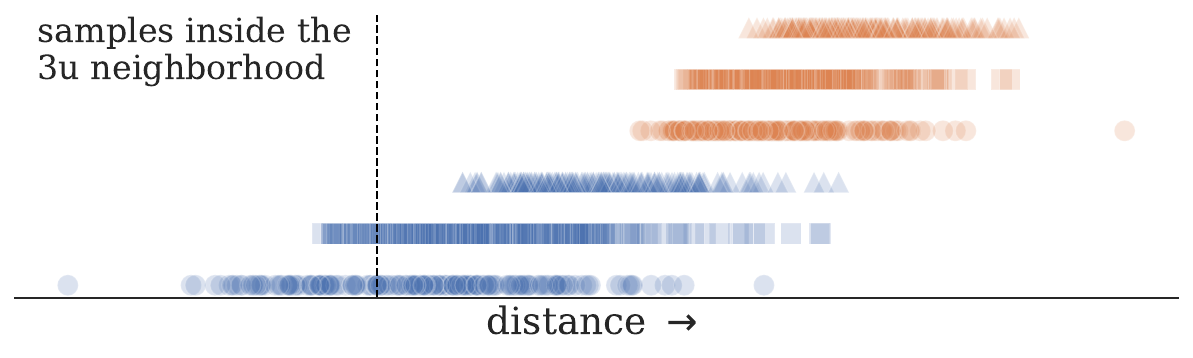}
	\caption[Optional caption]{High-dimensional distances to a random \tikzcircle{blue} point aggregated per label. Only samples to the left of the dashed vertical line (all from  cluster \tikzcircle{blue} or \tikzsquare{blue}) will exert an attractive force.}
	\label{fig:similarity_pij}
\end{figure}

\vspace{4pt}\pstart{The problem} The goal of ct-SNE is to bring secondary structure to the front by discounting certain points, i.e., increasing repulsive forces for points with the same label. With a fixed number of $3u$ nearest neighbors---that directly affect the placement of each point---ct-SNE can still only show new structure that reaches into this neighborhood. For the synthetic data, the cluster sizes are larger than $3u$, hence points with different labels are by definition outside the $3u$ neighborhood. In Figure \ref{fig:similarity_pij} we show the points that are part of $\mathcal{N}_{\tikzcircle{blue}}$. Since the neighbors all have the same \textbf{\textcolor{blue}{blue}} label, ct-SNE has no information on the similarity to \textbf{\textcolor{orange}{orange}} labeled points. The overlap between the \tikzcircle{blue} and \tikzsquare{orange} points in Figure \ref{fig:embeddings_intro}b occurs solely due to the decreased repulsive forces between differently-labeled samples. It is coincidental and \emph{wrong}, in the sense points from $\tikzcircle{orange}$ are closer to $\tikzcircle{blue}$, but this information is omitted and a wrong solution emerges (Figure~\ref{fig:embeddings_intro}b).

\begin{figure}[t]
	\centering
	\begin{subfigure}[b]{.45\textwidth}
		\includegraphics[height=40mm]{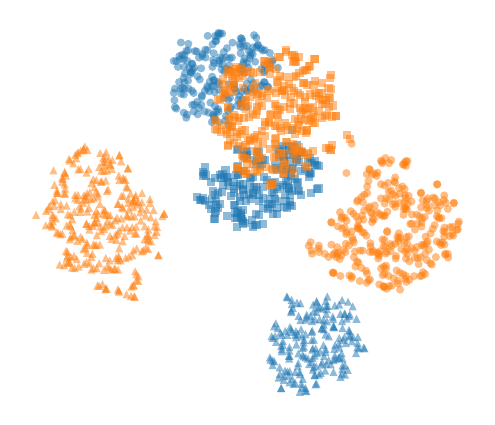}
		\caption{perplexity 30, $|\mathcal{N}| = n$\label{fig:ctsne_noapprox}}
	\end{subfigure}\hfill
	\begin{subfigure}[b]{.45\textwidth}
		\centering
		\includegraphics[height=40mm]{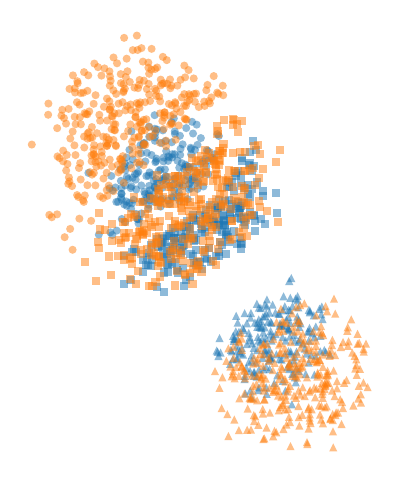}
		\caption{perplexity 300, $|\mathcal{N}| = 3*\text{perplexity}$\label{fig:ctsne_perp300}}
	\end{subfigure}\hfill
	\caption{Visualizations of ct-SNE embeddings ($\beta = \expnumber{1}{\text{-}12}$) with color label as prior knowledge. Not approximating the similarities in (a), the clusters still overlap arbitrarily as the attractive forces are too small. A higher perplexity (b) leads to correctly merged clusters that are not well separated. Changing $\beta$ does not affect the results.\label{fig:ctsne_solutions}}
\end{figure}

\vspace{4pt}\pstart{Alternative solutions} Increasing the neighborhood size does not lead to the desired results. We explain two ways that seem promising to circumvent the problem but do not work in practice. First, one could keep all pairwise HD similarities instead of approximating them. This results in non-zero attractive forces for differently-labeled points, but increases the complexity to compute these forces in every gradient update to $\mathcal{O}(n^2)$. In addition, the KL-divergence is asymmetric and weighs high $p_{ij}$ to be more important to match with the LD similarity than small similarities. We implemented this method and see in Figure \ref{fig:ctsne_noapprox} that the embedding on the synthetic data has barely changed. We presume the HD similarities are too small to have an effect on the embedding. 

The second idea is to increase the perplexity. A higher perplexity will increase the neighborhood size by definition and differently-labeled neighbors might get assigned a higher similarity than with the first solution (and smaller perplexity). For large datasets with few class labels to be factored out, this might still be impractical, because it could be necessary to use a perplexity of $n/2$ to have sufficiently high attractive forces\footnote{Assuming the distances between differently-labeled samples are larger than between same-labeled samples. A perplexity of $\frac{n}{2}$ would only assign same-labeled points a high similarity.}. What is even more unfavorable is the loss of \textit{locality} that goes hand in hand with a higher perplexity and stands in opposition with the original idea of t-SNE to preserve local neighborhoods. Figure  \ref{fig:ctsne_perp300} shows that on the synthetic data a high perplexity indeed leads to better preservation of the similarities between differently-labeled points, but locality is lost and instead two clusters emerge, instead of the expected three clusters.

\section{Revised Conditional t-SNE}
In this section we argue how an adjusted approximation and a different formulation of ct-SNE might help retain important neighborhood information when factoring out prior knowledge. We propose two changes to ct-SNE to provide enough structural information about differently-labeled nearest neighbors and still discount the similarity to same-labeled points.

\vspace{4pt}\pstart{Expanding the set of nearest neighbors} First, we search for nearest neighbors separately for same and differently-labeled points. We use $\mathcal{N}_{i, \delta_{ij}=1}$ as the set of $1.5u$ nearest neighbors with the same class label as $i$ and $\mathcal{N}_{i, \delta_{ij}=0}$ denotes the set with $1.5u$ differently-labeled nearest neighbors. This does not add runtime to the gradient updates (when still using $3u$ neighbors in total) but requires to build and search in separate nearest-neighbor data structures (e.g., vantage-point trees \citep{yianilos1993data}, ANNOY \citep{annoy}) for each label. As we saw in Figure \ref{fig:ctsne_noapprox}, this change alone will not be sufficient.

\begin{figure}[!b]
	\begin{subfigure}{.48\textwidth}
		\includegraphics[width=\textwidth]{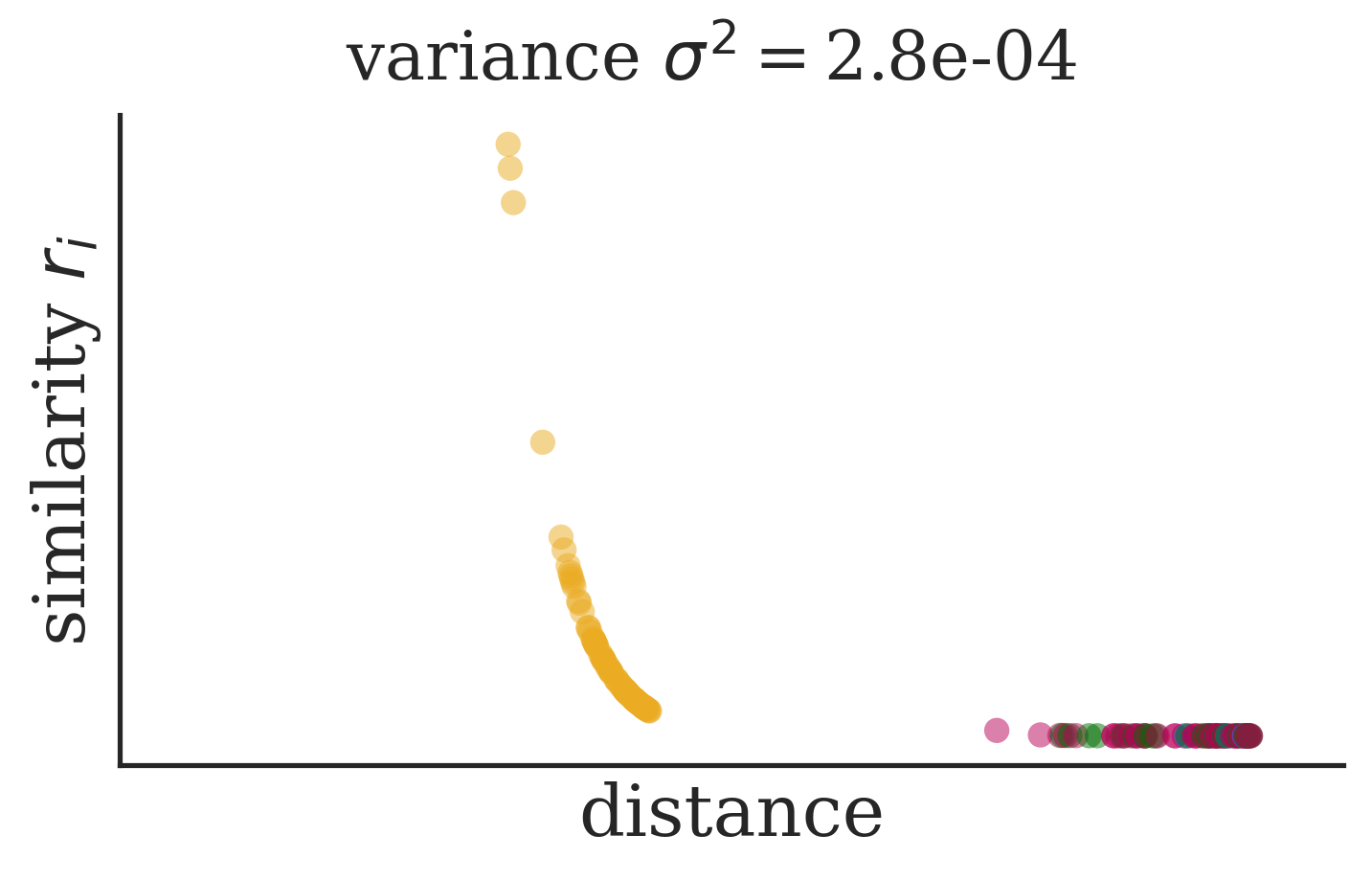}
		\caption{}
		\label{fig:entropy_a}
	\end{subfigure}\hfill
	\begin{subfigure}{.48\textwidth}
		\includegraphics[width=\textwidth]{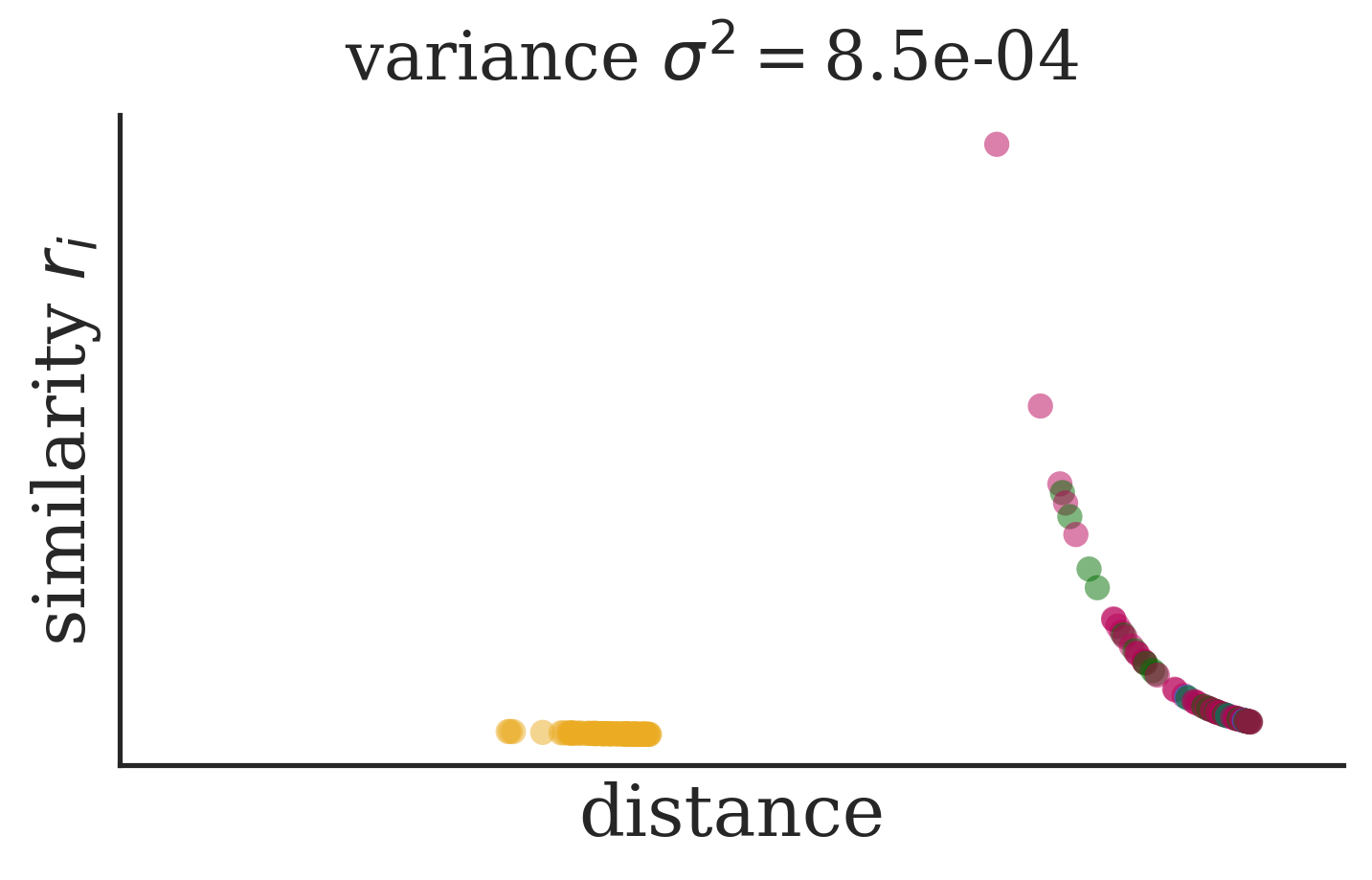}
		\caption{}
		\label{fig:entropy_b}
	\end{subfigure}
	\caption{Two similarity distributions $r_{i}$ with $\beta = \expnumber{1}{\text{-}4}$ where a different variance leads to the same perplexity of $50$. The distances are computed for cell $i=12823$ (indrop, beta) of the pancreas dataset and the colors correspond to the technology label.}
	\label{fig:entropy_variance}
\end{figure}

\vspace{4pt}\pstart{Condition the high-dimensional similarities} The second change is to condition the HD instead of the LD similarities. This will affect the attractive forces, in contrast to the repulsive forces in ct-SNE. We define
\begin{align*}
	r_{j|i} = P_{p_i}(j|\Delta) &= \frac{P_{p_i}(\Delta|j)\cdot p_{j|i}}{P_{p_i}(\Delta)}\,,\\
\end{align*}
where $P_{p_i} (\Delta|j) = \frac{\prod_{l}n_l!}{n!}\beta^{\delta_{ij}} \alpha^{1-\delta_{ij}}$ is defined as in ct-SNE but we flipped the parameters\footnote{We use $\alpha$ and $\beta$ instead of $\alpha'$ and $\beta'$ as in the original paper \cite{kang2021conditional}, and thus need to normalize with the number of distinct label assignments.}. The notation $P_{p_i}$ denotes that we compute the similarity distribution for a fixed sample $i$ and the corresponding values of $p_{<\cdot>|i}$. The marginal probability is also defined for each $i$ separately as
\begin{align*}
	P_{p_i}(\Delta) &= \sum_{k\neq i} P_{p_i}(\Delta|k) \cdot p_{k|i} = \beta \sum_{k\neq i: \delta_{ik} = 1} p_{k|i} + \alpha \sum_{k\neq i: \delta_{ik} = 0} p_{k|i}\,.
\end{align*}
The HD similarities for all samples in the neighborhood are
\begin{align*}
	r_{j|i} &= 
	\begin{cases}
		\frac{\beta p_{j|i}}{\beta\sum_{k\neq i:\delta_{ik} = 1} p_{k|i} + \alpha\sum_{k\neq i: \delta_{ik} = 0} p_{k|i}} \qquad &\text{if } j \in \mathcal{N}_{i, \delta_{ij}=1}\\
		\frac{\alpha p_{j|i}}{\beta\sum_{k\neq i:\delta_{ik} = 1} p_{k|i} + \alpha\sum_{k\neq i: \delta_{ik} = 0} p_{k|i}} \qquad &\text{if } j \in \mathcal{N}_{i, \delta_{ij}=0}\,,
	\end{cases}
\end{align*}
where the relation between $\alpha$ and $\beta$ is the same as in ct-SNE, and we will use $\beta < \alpha$ to decrease the similarity of same-labeled data points. Finally, the similarities are symmetrized as before $r_{ij} = (r_{j|i} + r_{i|j})/2n$ and the loss $\kl{r}{q}$ measures the KL-divergence between the conditioned HD similarities and the LD similarities. Since the input similarities do not depend on the embeddings $y_i$, the gradient is the same as in t-SNE with $r_{ij}$ instead of $p_{ij}$. This allows us to integrate our changes into FIt-SNE \cite{linderman2019fast} offering a fast interpolation-based acceleration of the gradient computation. Our code is available at \href{https://github.com/aida-ugent/revised-conditional-t-SNE.git}{github.com/aida-ugent/revised-conditional-t-SNE}.

\vspace{4mm}\pstart{Estimating the point-wise variance} This new formulation of adjusting the HD similarities raises the question whether the point-wise variance of the Gaussian kernel should be computed using $r_{i}$ or $p_{i}$. On $r_i$, the binary search for the variance satisfying the user-defined perplexity is not well-defined as the perplexity is not monotonously increasing with the variance. Thus, two or more possible solutions exist that can have opposite characteristics as shown in Figure \ref{fig:entropy_variance}. 
The other option is to first estimate the variance on $p_{ij}$ and then change the similarities with $\alpha$ or $\beta$. However, the effective perplexity of $r_{ij}$ might differ from the specified perplexity defined by the user.

\section{Evaluation}
\label{sec:evaluation}
To compare \ourmethod with ct-SNE we provide experimental results on a synthetic and two biological datasets. We first describe the evaluation setup including the chosen quality measures and then describe the results. Embeddings of \ourmethod with variance estimation on $p_i$ and experimental results on the second biological dataset can be found in the supplement.

\subsection{Setup}
We first provide the characteristics of the datasets and then define the evaluation measures. All experiments were run on a laptop with Intel® Core™ i7-10850H CPU @ 2.70GHz with 16GB RAM.

\vspace{4mm}\pstart{Datasets}
\begin{description}
	\item[Synthetic data] Each point in this $n=1500, d=10$ dataset belongs to one of two clusters in dimensions 1-4 and one of three clusters in dimensions 5-6. 
	The cluster centers are sampled from $\mathcal{N}(0, 25)$ and $\mathcal{N}(0, 1)$ respectively. For each point, we add noise from $\mathcal{N}(0, 0.01)$ to the cluster centers and append four dimensions of noise from $\mathcal{N}(0,1)$. The clusters in dim 5-6 are of equal size, while 600 points belong to \textcolor{blue}{blue} and 900 to the \textcolor{orange}{orange} cluster. We provide the cluster labels of dim 1-4 as prior knowledge to ct-SNE and expect the embedding to show the structure implanted in dimensions 5-6. 
	\item[Pancreas data \cite{satija2019panc8}] is a widely-used single cell RNAseq dataset ($n=14890, d=34363$) to benchmark data integration methods. It contains gene counts of human pancreatic islets cells from 8 sources sequenced with 5 different technologies\nolinebreak---SMARTSeq2 (2394), Fluidigm C1 (638), CelSeq (1004), CelSeq2 (2285), and inDrops (8569). We provide the technology labels as prior knowledge to merge cells from different technologies together and expect a grouping according to the 13 celltypes. We followed the standard preprocessing steps for single-cell RNA datasets including the selection of 2000 highly-variable genes, normalization, standardization, and PCA to retain 50 principal components. 
\end{description}

\vspace{4mm}\pstart{Evaluation measures}
To compare the embeddings quantitatively, we compute a normalized HD and LD neighborhood overlap score \citep{lee2009quality,lee2015multi} and the degree of label mixing with the \textit{Laplacian score} that was also used to evaluate ct-SNE. For both measures we use a fixed neighborhood size equal to the perplexity which is 30 for the synthetic and 50 for the pancreas dataset. For the pancreas data, we compute both measures on a random subset of 5\% of the data.
\begin{description}
	
	\item[$\mathbf{R_{NX}}$ neighborhood preservation] %
	measures the normalized agreement of HD and LD neighborhoods as proposed by \citet{lee2009quality}. Denoting the $k$-sized HD and LD neighborhoods of data point $i$ as $v_i^k$ and $n_i^k$, the average neighborhood overlap rate is defined as
	$$Q_{NX}(k) = \frac{1}{kn}\sum_{i = 1}^n |v_i^k \cap n_i^k|\,.$$
	Since a random embedding would yield a score of $\mathbb{E}[Q_{NX}(k)] = \frac{k}{n-1}$, these values are scaled to $R_{NX}(k) = \frac{(n-1)Q_{NX}(k) - k}{n-1-k} \in [0,1]$, measuring the improvement over a random embedding. We adjust this measure to reflect the idea of factoring out class label information. Given a set of LD neighbors $n_i^k$, we ensure that $v_i^k$ contains equally many points with the same (and different) label, i.e., $|\{j \,|\,j\in v_i^k, \delta_{ij}=1\}| =|\{j\,|\,j\in n_i^k,\delta_{ij}=1\}|$. The distribution of same and differently labeled neighbors is determined by the embedding and differs per point.
	
	\item[Laplacian scores] proposed by \citet{kang2021conditional} measure the fraction of LD nearest neighbors with a different label. It can be compared to a baseline with random label assignment, where the expected Laplacian score is $\sum_{l \in L} \frac{n_l(n-n_l)}{n(n-1)}$ where label $l\in L$ has $n_l$ samples. When factoring out structure encoded by a labeling of the data (e.g. dim 1-4 labels for the synthetic data or the technology feature for pancreas), we expect an increase of the Laplacian score evaluated on the same label. An increase of the Laplacian evaluated on a different class label is not necessarily desirable. 

\end{description}

\begin{figure}[t]
	\begin{subfigure}{\textwidth}
		\includegraphics[width=\textwidth]{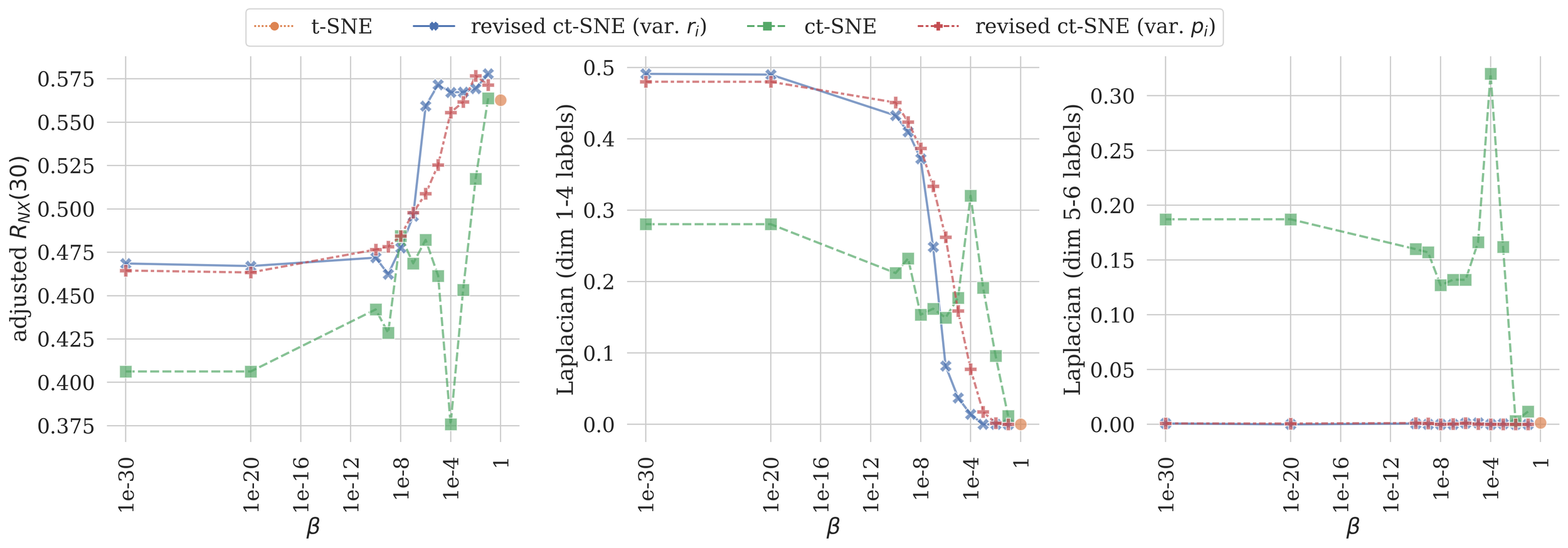}
		\caption{\rnx and Laplacian scores for different values of $\beta$ evaluated on neighborhoods of size $k=30$. The t-SNE scores are depicted at $\beta = 1$.}
		\label{fig:synthetic_evaluation_scores}
	\end{subfigure}
	\begin{subfigure}{.45\textwidth}
		\includegraphics[width=\textwidth]{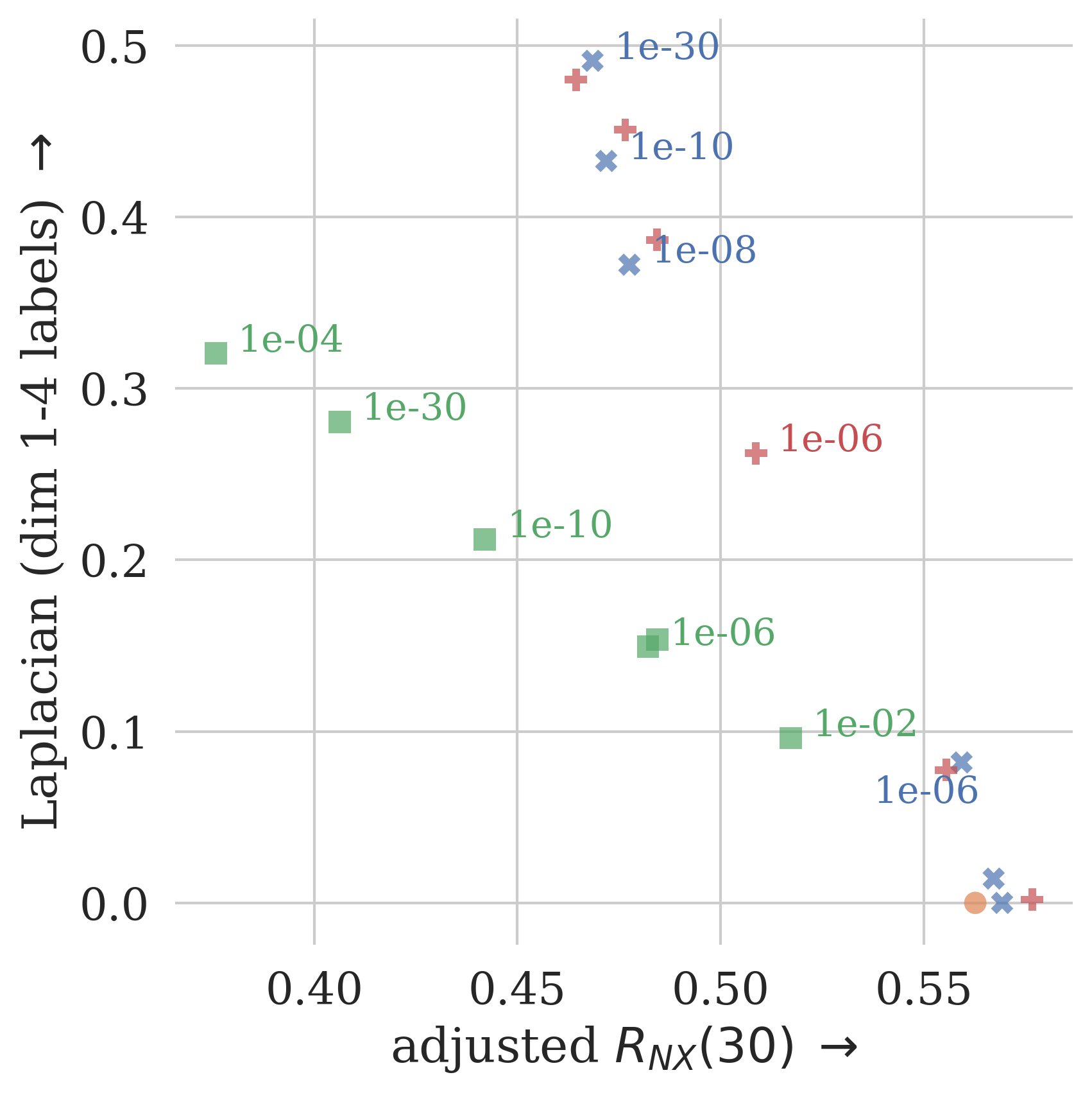}
		\caption{Trade-off between \rnx and Laplacian scores based on the dim1-4 labels.}
		\label{fig:synthetic_rnx_laplacian_scores}
	\end{subfigure}\hfill
	\begin{subfigure}{.45\textwidth}
		\includegraphics[width=\textwidth]{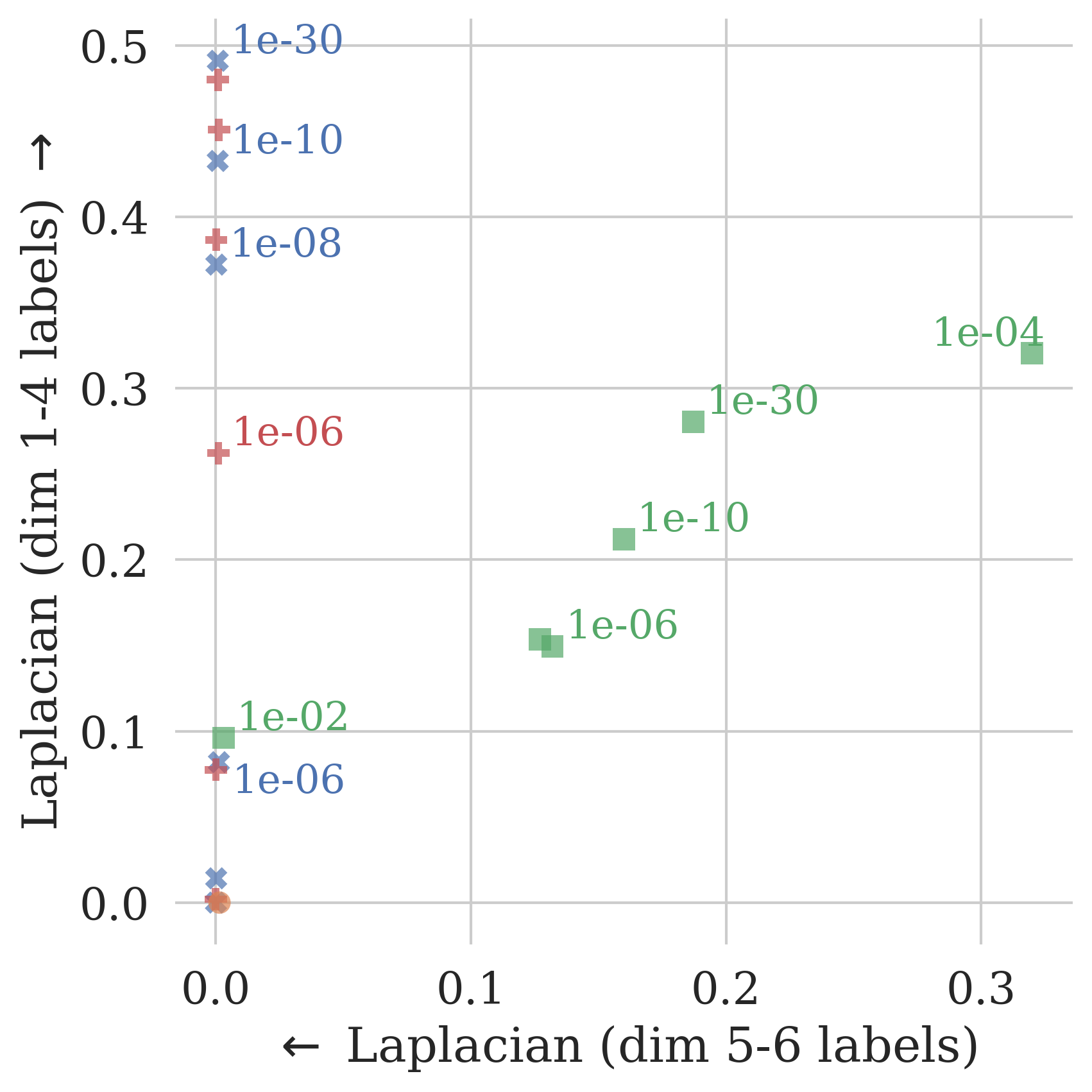}
		\caption{Trade-off between Laplacian scores.}
		\label{fig:synthetic_laplacian_scores}
	\end{subfigure}
	\caption{Evaluation results for the synthetic dataset. In \ref{fig:synthetic_evaluation_scores} we show the neighborhood agreement and Laplacian for varying $\beta$, while \ref{fig:synthetic_rnx_laplacian_scores} and \ref{fig:synthetic_laplacian_scores} allow to compare two of the measures for a subset of all $\beta$ values.}
	\label{fig:syntheti_results}
\end{figure}

\subsection{Results}
We compare embeddings of t-SNE with embeddings by ct-SNE and \ourmethod for different values of $\beta$, and using variance estimation with either $r_i$ or $p_i$.

\vspace{4mm}\pstart{Synthetic data} 
To embed the synthetic dataset, we use a perplexity of $u=30$, $\theta = 0.2$, and $750$ epochs. This took about 10s for all methods. The \rnx and Laplacian scores are shown in Figure \ref{fig:syntheti_results} and in Figure \ref{fig:embeddings_intro} we show embeddings of ct-SNE with $\beta=\expnumber{1}{\text{-}4}$ and \ourmethod with $\beta=\expnumber{1}{\text{-}20}$ as they score highest on the Laplacian (dim 1-4). The t-SNE embedding has a Laplacian score of 0 since the 30 nearest neighbors have the same labels (same color and shape) as we can visually confirm in Figure \ref{fig:embeddings_intro}a. 
\Ourmethod converges to a Laplacian score (dim 1-4) equivalent to a random embedding. We conclude that the structure captured by the labels in dimensions 1-4 has successfully been factored out in the embedding. The embedding by ct-SNE scores lower on the Laplacian (dim 1-4 labels) but higher when using the labels in dimensions 5-6 ($\emptycirc, \triangle, \square$). This indicates that not only the imposed structure in dimensions 1-4 but also from dimensions 5-6 has been erroneously removed. The HD neighborhoods for a fixed level of the Laplacian (dim 1-4) are more accurately preserved by \ourmethod as shown in in Figure \ref{fig:synthetic_rnx_laplacian_scores}.

\begin{figure}[tp]
	\centering
	\begin{subfigure}{\textwidth}
		\includegraphics[width=\textwidth]{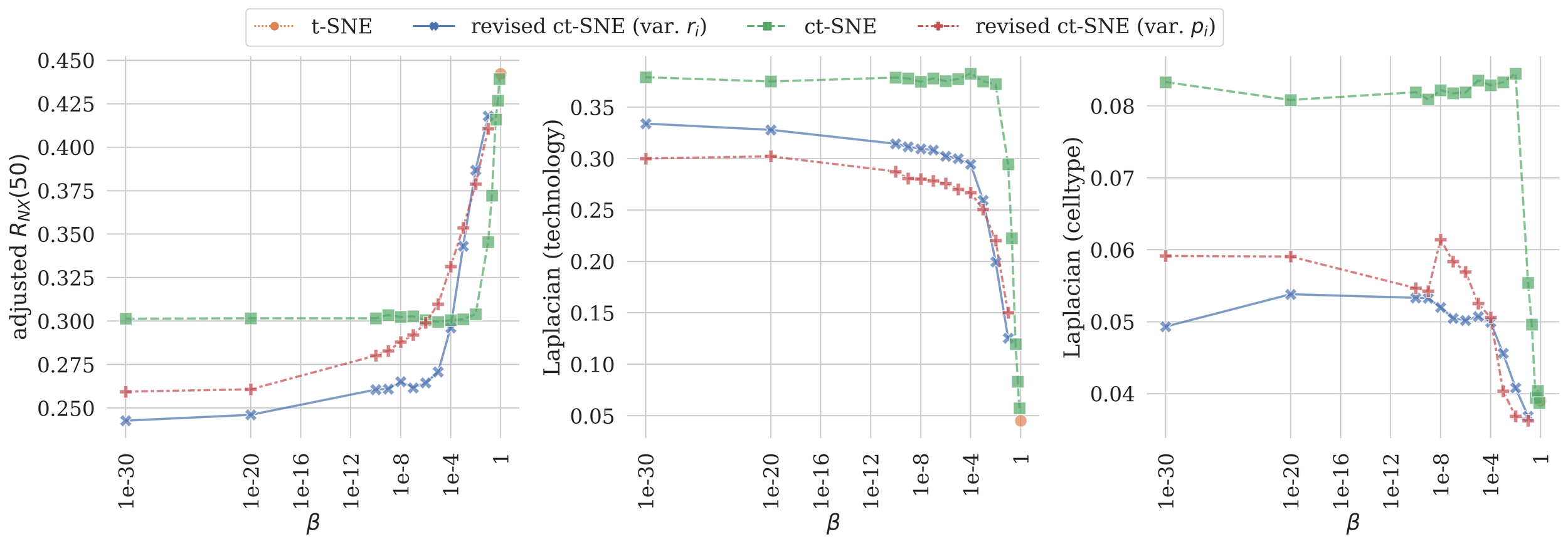}
		\caption{\rnx and Laplacian scores for different values of $\beta$ evaluated on neighborhoods of size $k=50$. The t-SNE scores are depicted at $\beta = 1$.}
		\label{fig:panc_evaluation_scores}
	\end{subfigure}\hfill
	\begin{subfigure}{\textwidth}
		\includegraphics[width=\textwidth]{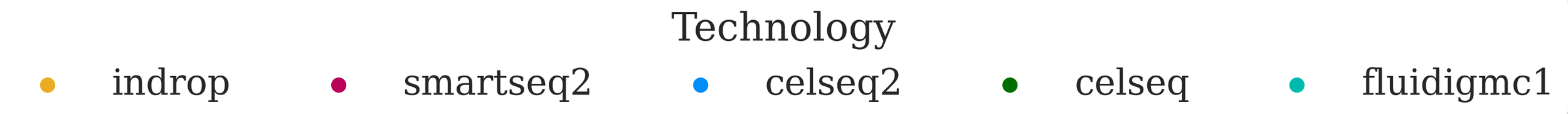}
	\end{subfigure}
	\begin{subfigure}[b]{.32\textwidth}
		\includegraphics[width=\textwidth]{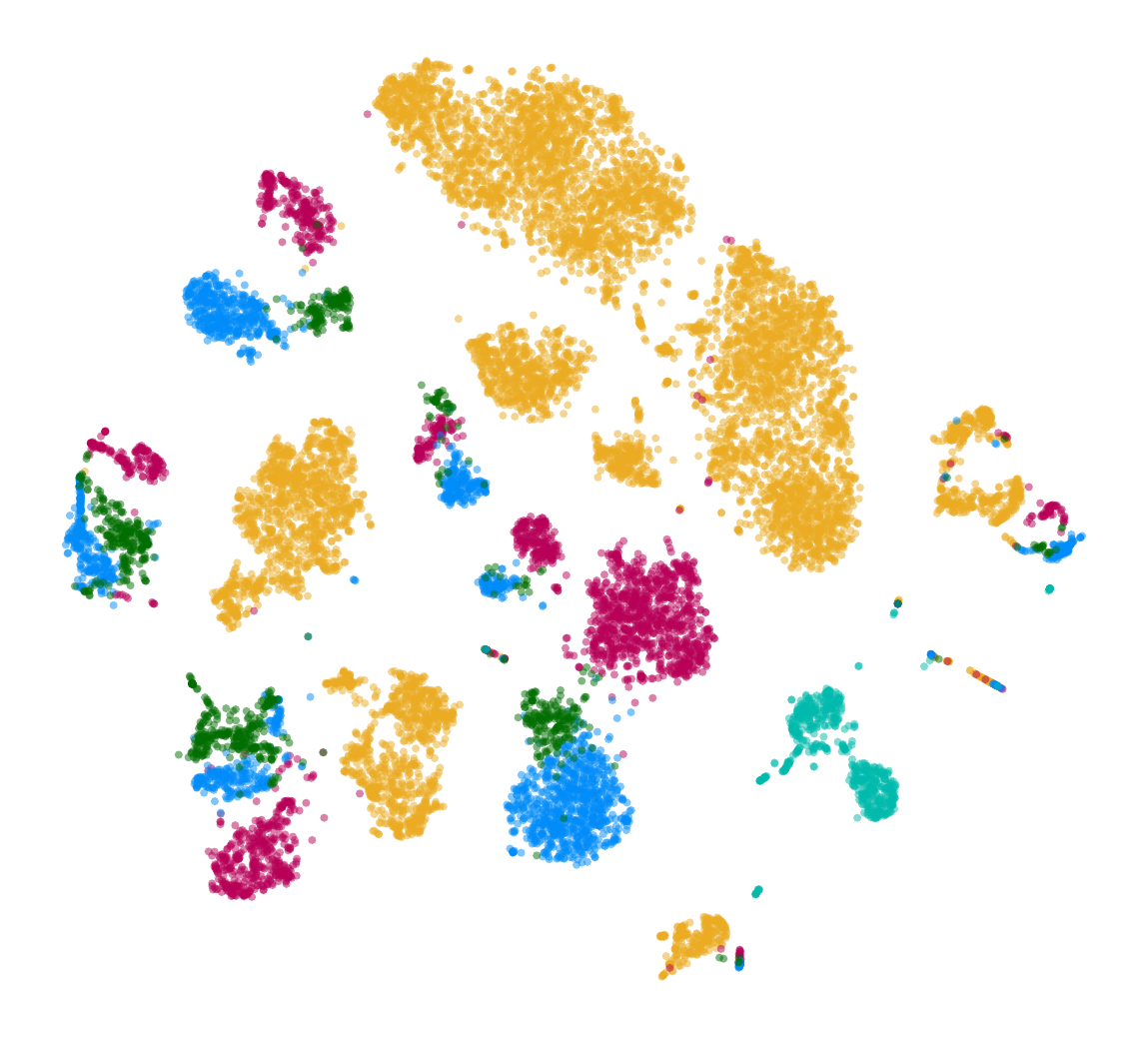}
		\caption{t-SNE}
		\label{fig:panc_tsne}
	\end{subfigure}\hfill
	\begin{subfigure}[b]{.28\textwidth}
		\includegraphics[width=\textwidth]{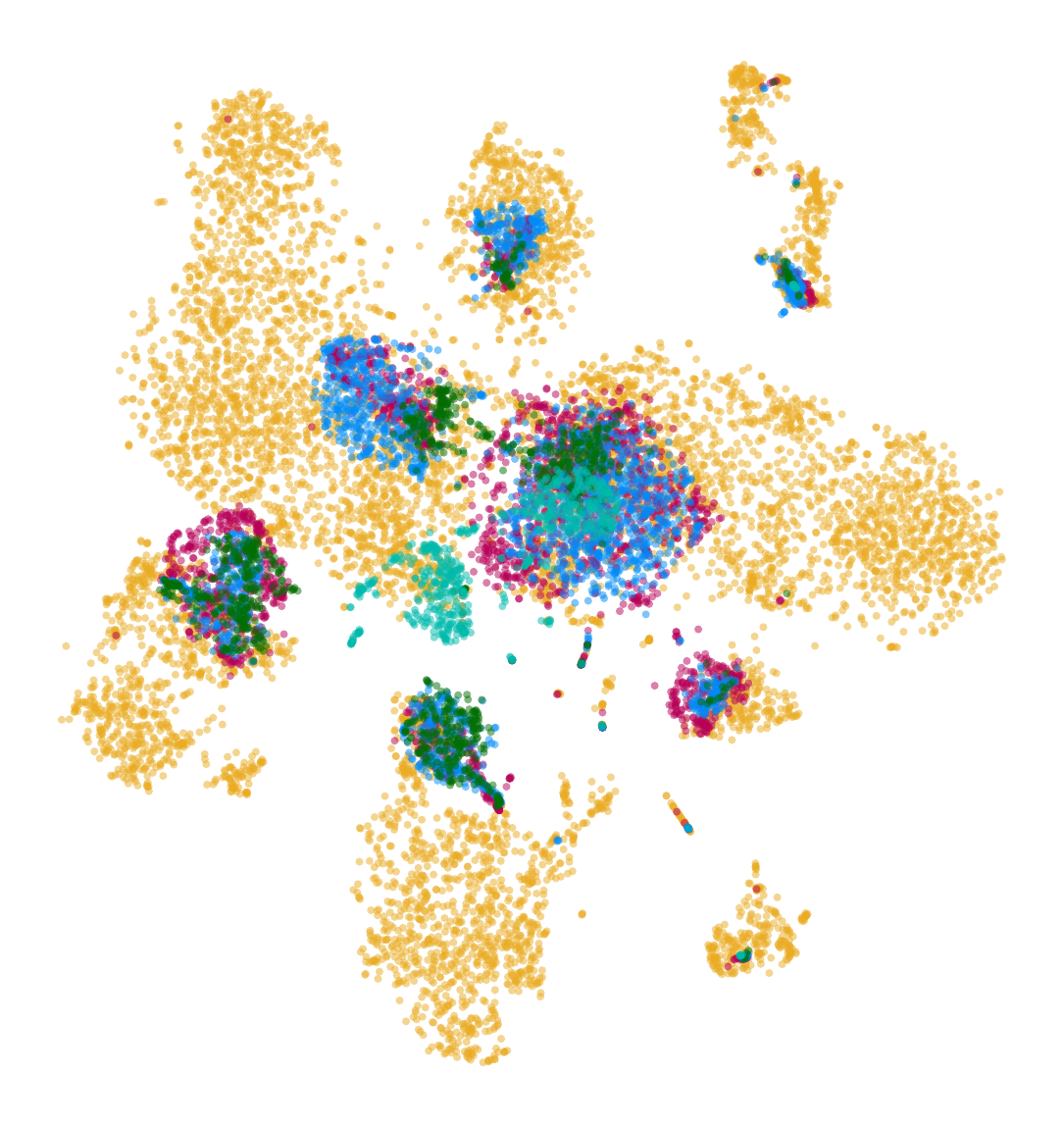}
		\caption{ct-SNE}
		\label{fig:panc_ctsne}
	\end{subfigure}\hfill
	\begin{subfigure}[b]{.36\textwidth}
		\includegraphics[width=\textwidth]{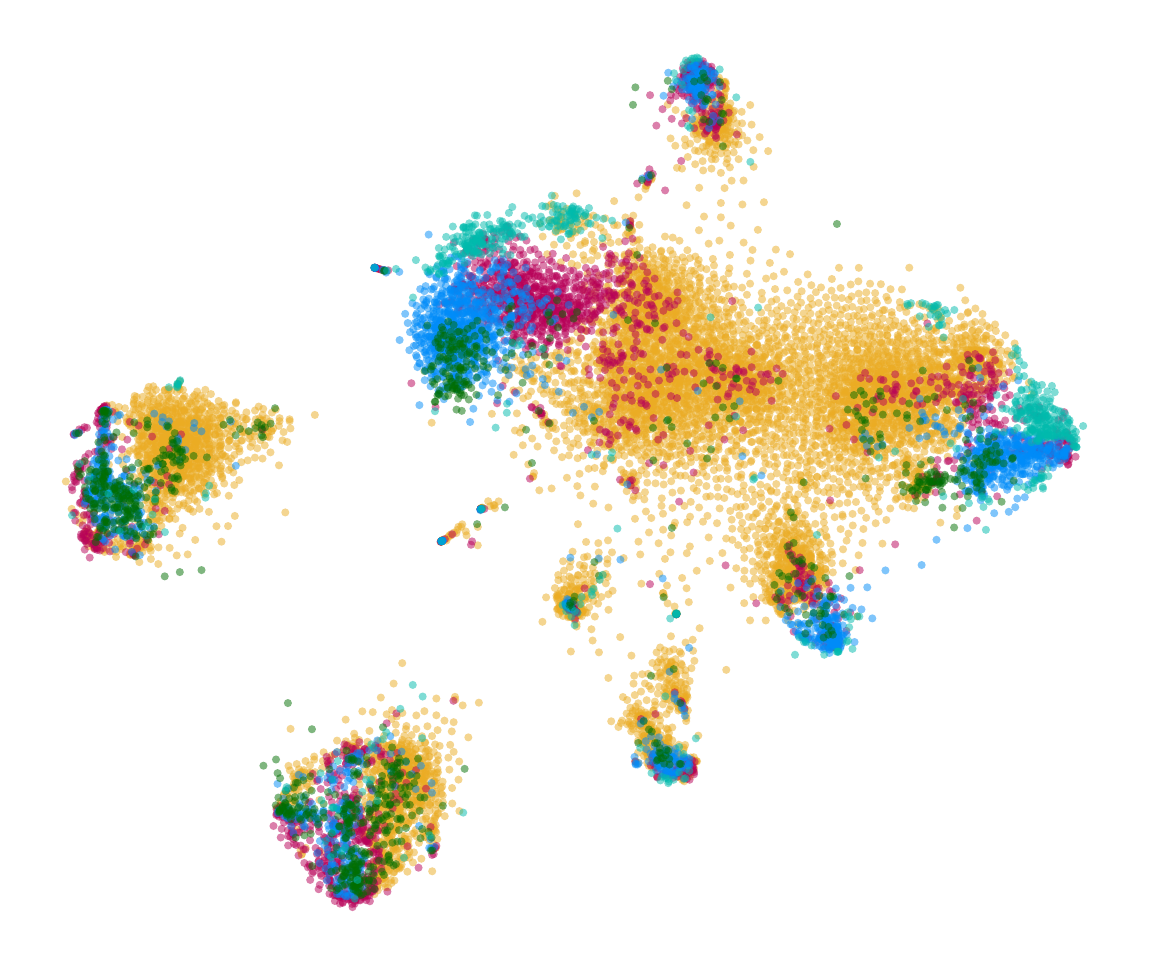}
		\caption{\ourmethod (var. $r_i$)}
		\label{fig:panc_fastctsne}
	\end{subfigure}\hfill
	\vspace{.3cm}
	\begin{subfigure}{\textwidth}
		\includegraphics[width=\textwidth]{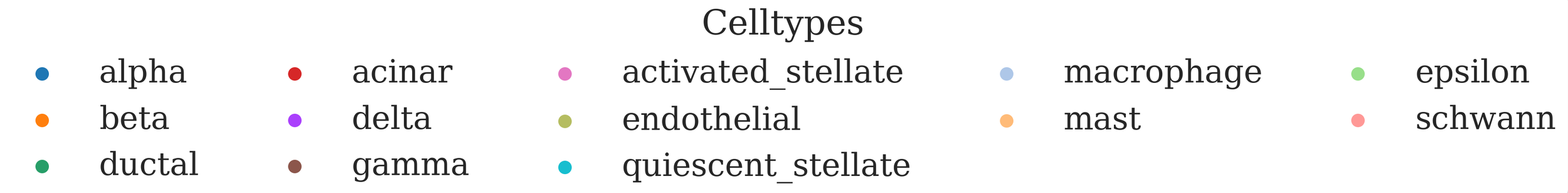}
	\end{subfigure}
	\begin{subfigure}[b]{.32\textwidth}
		\includegraphics[width=\textwidth]{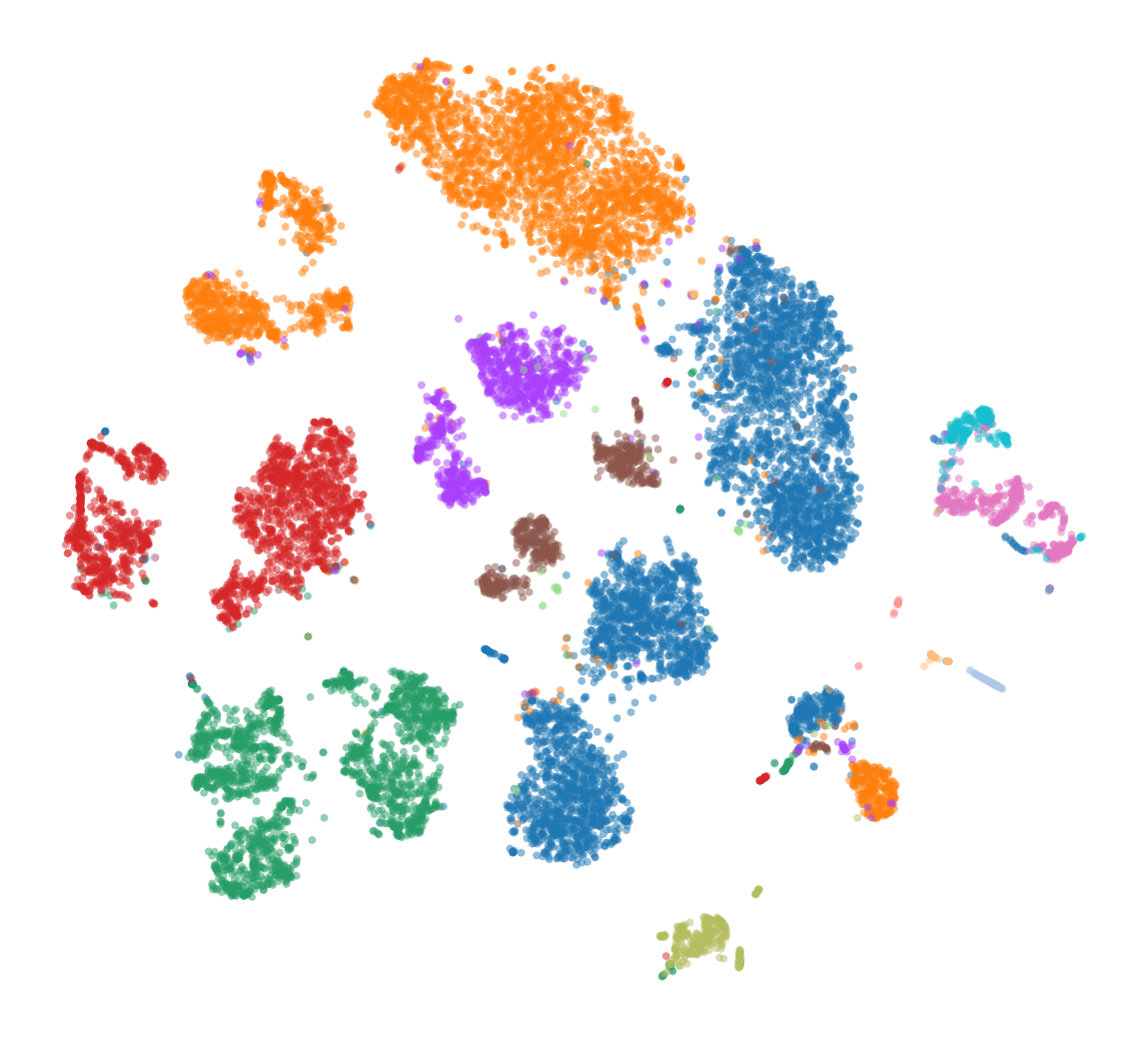}
		\caption{t-SNE}
		\label{fig:panc_tsne_celltypes}
	\end{subfigure}\hfill
	\begin{subfigure}[b]{.28\textwidth}
		\includegraphics[width=\textwidth]{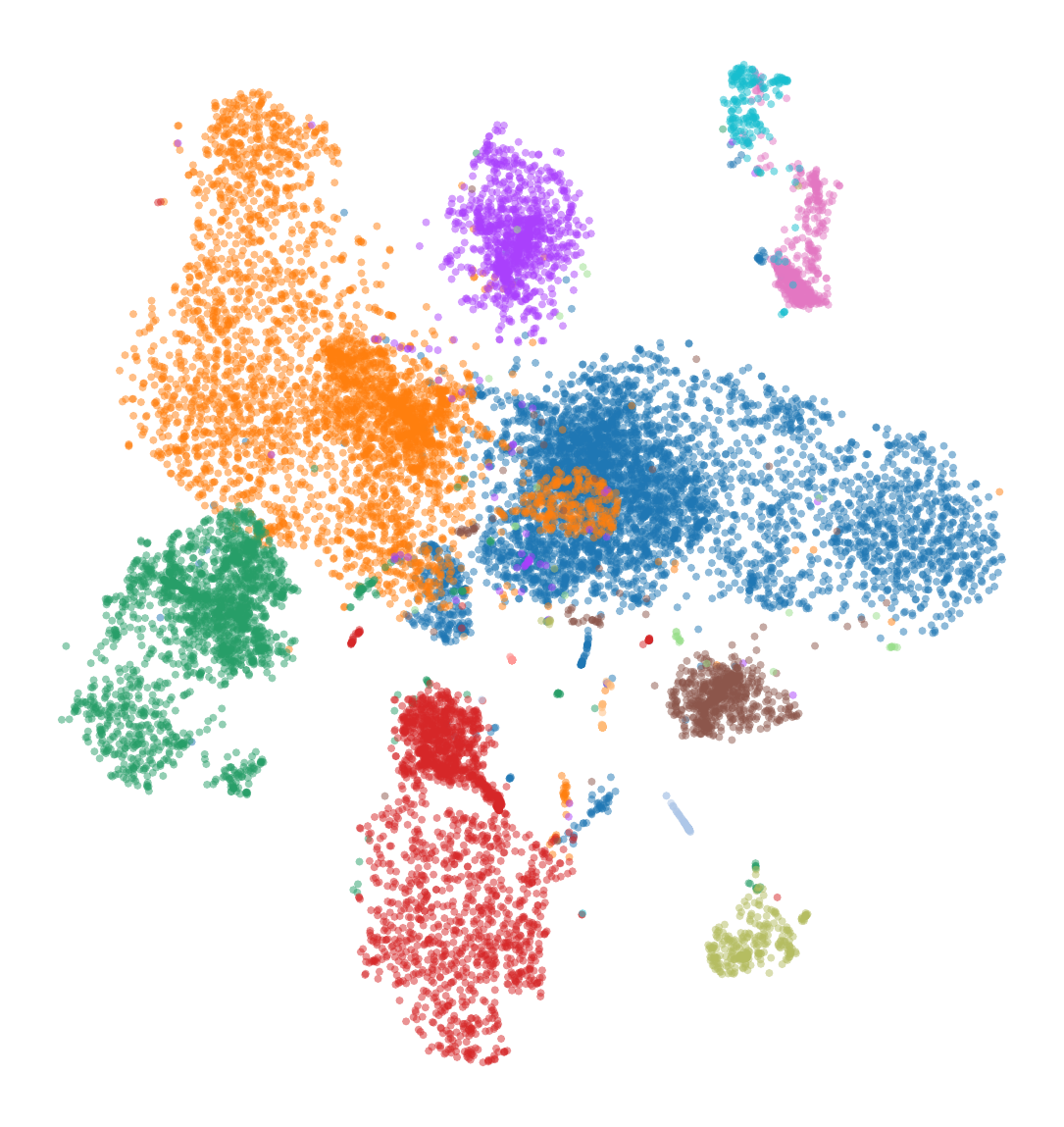}
		\caption{ct-SNE}
		\label{fig:panc_ctsne_celltypes}
	\end{subfigure}\hfill
	\begin{subfigure}[b]{.36\textwidth}
		\includegraphics[width=\textwidth]{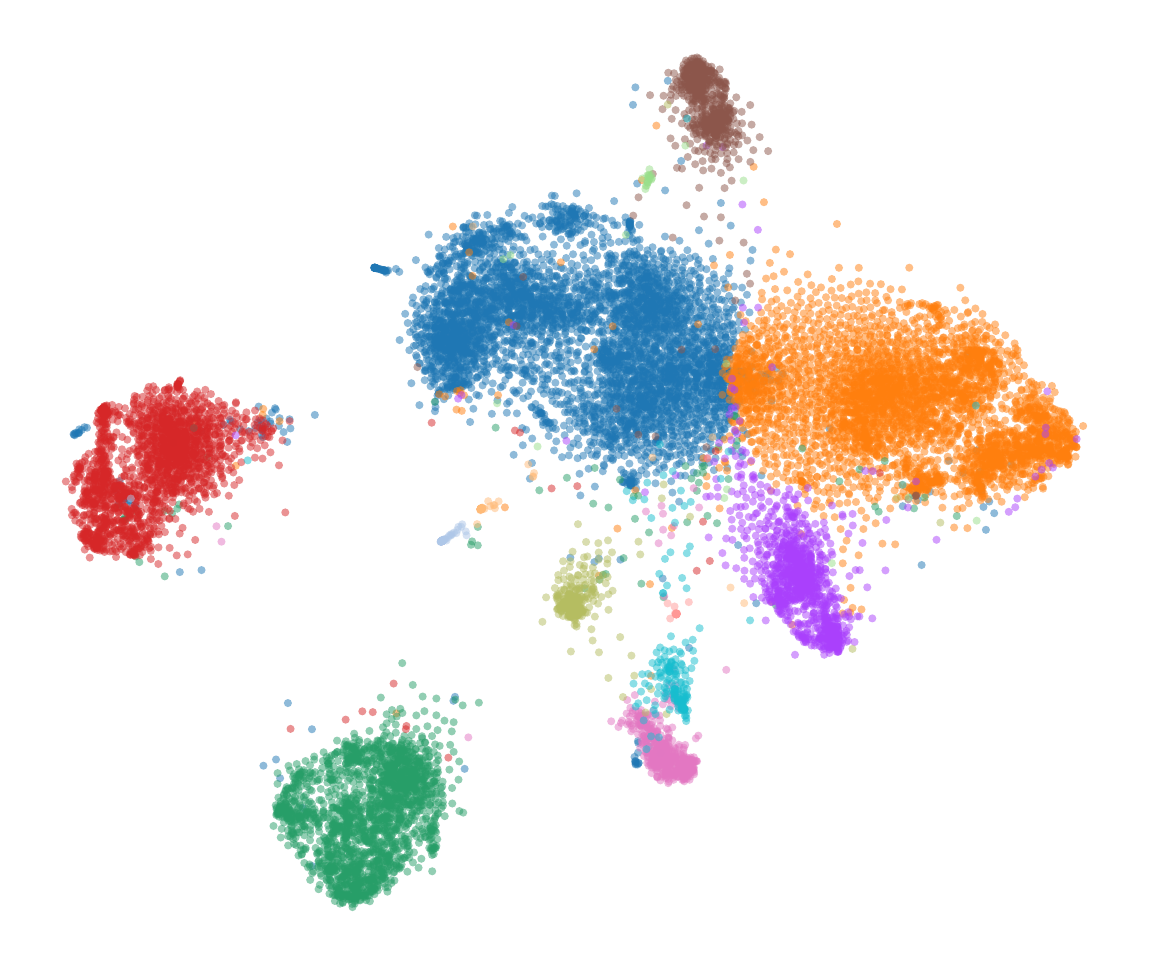}
		\caption{\ourmethod (var. $r_i$)}
		\label{fig:panc_fastctsne_celltypes}
	\end{subfigure}
	\caption{Visualizations and Laplacian scores of pancreas data embeddings where the technology labels are provided as prior information to ct-SNE and \ourmethod with $\beta = \expnumber{1}{\text{-}30}$. Cell coloring by technology (\ref{fig:panc_tsne})-(\ref{fig:panc_fastctsne}) and cell type (\ref{fig:panc_tsne_celltypes})-(\ref{fig:panc_fastctsne_celltypes}).}
	\label{fig:panc_embeddings}
\end{figure}

\begin{figure}
	\begin{subfigure}{\textwidth}
		\includegraphics[width=\textwidth]{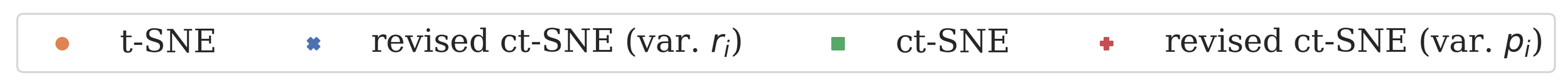}
	\end{subfigure}
	\begin{subfigure}{.45\textwidth}
		\includegraphics[width=\textwidth]{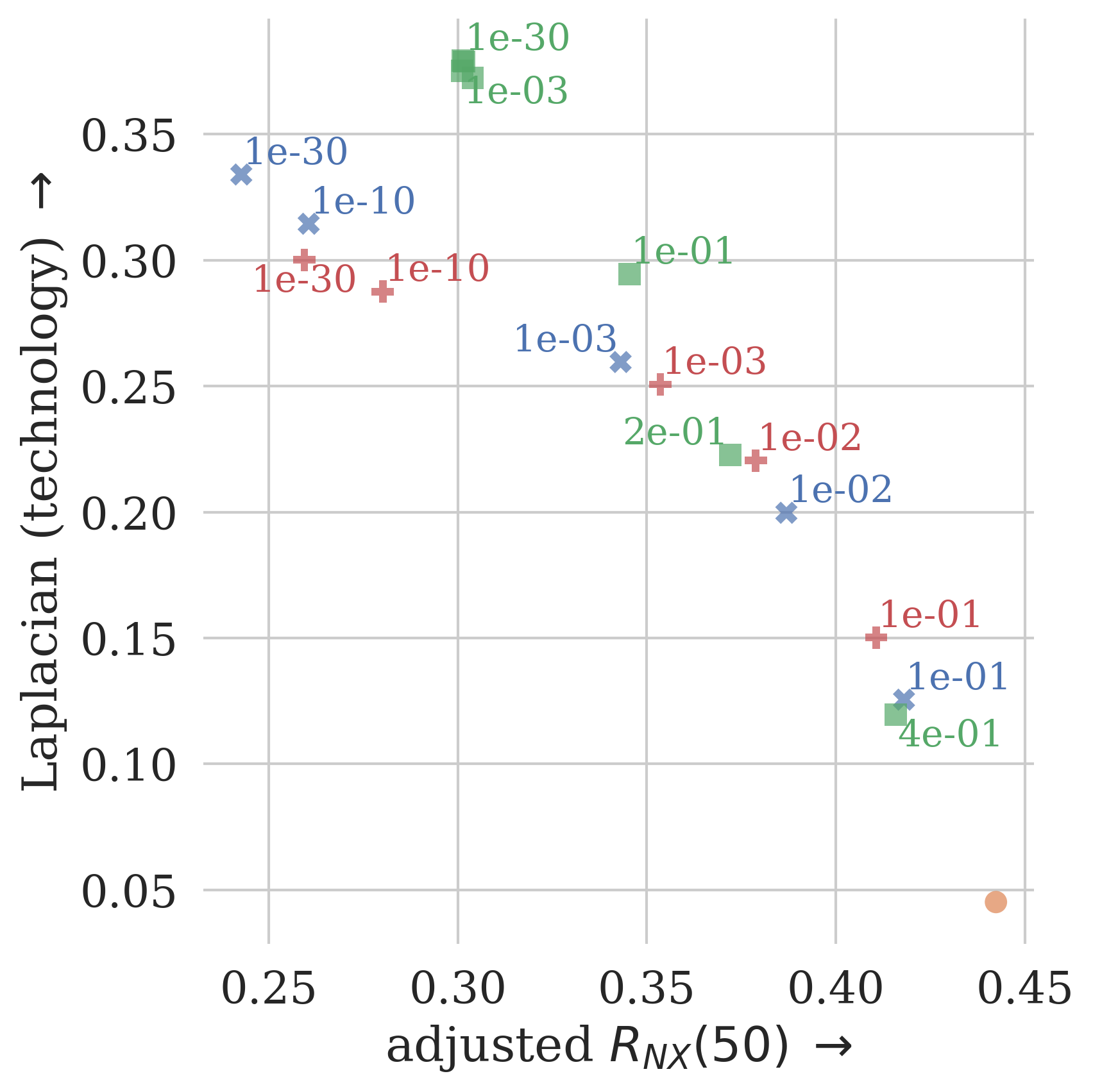}
		\caption{}
		\label{fig:panc_rnx_tradeoff}
	\end{subfigure}\hfill
	\begin{subfigure}{.45\textwidth}
		\includegraphics[width=\textwidth]{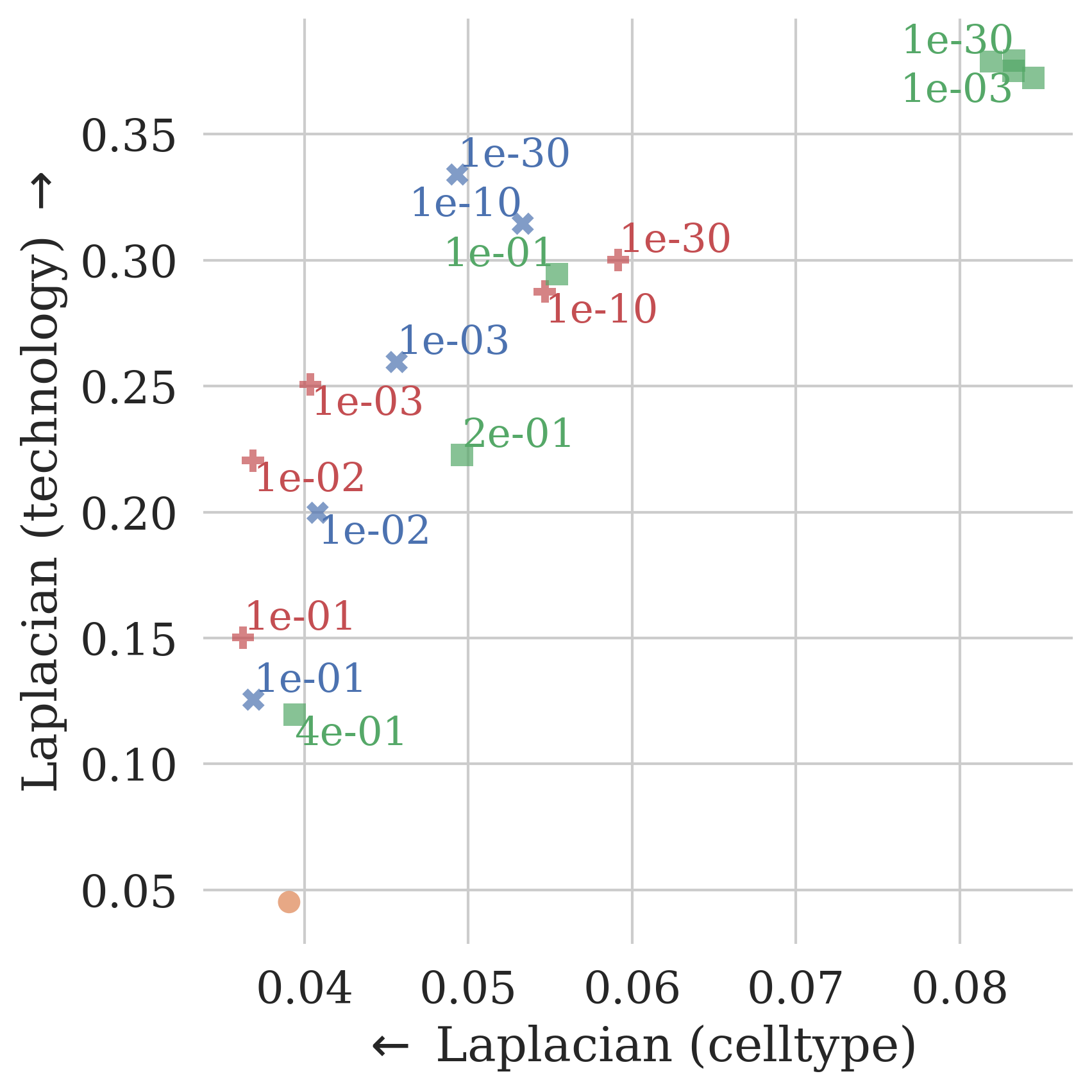}
		\caption{}
		\label{fig:panc_laplacian_tradeoff}
	\end{subfigure}
	\caption{Evaluation scores for embeddings of the pancreas data. Trade-offs between \rnx and Laplacian in (a) and Laplacian technology versus celltype in (b).}
\end{figure}

\vspace{4mm}\pstart{Pancreas data}
To embed the pancreas data we use a perplexity of $50$, $\theta = 0.5$, and $1000$ iterations and show the t-SNE and conditional embeddings with $\beta = \expnumber{1}{\text{-}30}$ in Figure \ref{fig:panc_embeddings}. The runtime of ct-SNE for the pancreas dataset is $168s$ compared to $28s$ for \ourmethod.
The evaluation results depicted in Figure \ref{fig:panc_evaluation_scores} show that the neighborhood agreement $R_{NX}(50)$ drops significantly from 0.44 for t-SNE to 0.30 (ct-SNE) and 0.24 (\ourmethod) with $\beta = \expnumber{1}{\text{-}30}$. 
Conditional t-SNE and \ourmethod both converge to a Laplacian (technology) score that is lower than the score for a random mixing of labels (0.61). The Laplacian for the celltype labels however is high for ct-SNE, indicating a mix of different celltypes in the local neighborhoods. Indeed, visualization of the trade-offs (Figures~\ref{fig:panc_rnx_tradeoff} and \ref{fig:panc_laplacian_tradeoff}) learns us that ct-SNE manages to retain $R_{NX}$ better, while mixing the cells from different technologies, whereas revised ct-SNE leads to better trade-offs between the Laplacian scores (less mixing between celltypes, while mixing samples from different batches).

We also notice that $\beta > \expnumber{1}{\text{-}4}$ is sufficient for ct-SNE while the Laplacian (technology) for \ourmethod only plateaus for smaller values of $\beta$. We speculate that \ourmethod requires smaller values as we change Gaussian HD similarities instead of values from a fat-tailed t-distribution. The trade-offs visualized in Figures \ref{fig:panc_rnx_tradeoff} and \ref{fig:panc_laplacian_tradeoff} suggest that $\beta = 0.1$ for ct-SNE and $\beta = \expnumber{1}{\text{-}4}$ for \ourmethod might be suitable starting points that can be adjusted in both directions. Finally, the differences in evaluation scores between point-wise variance estimation using $r_i$ and $p_i$ are small and inconclusive.

\section{Discussion}
The experiments showed that \ourmethod outperformed ct-SNE on the synthetic dataset. 
On the pancreas dataset, ct-SNE achieved a better trade-off between neighborhood preservation and mixing cells from different technologies, but at the same time in a higher fraction of neighbors with a different celltype.
\Ourmethod benefits from the FIt-SNE implementation, leading to a faster runtime.

First, we explain the different results of ct-SNE on the two datasets. The original implementation of ct-SNE can suppress a grouping of data points according to given labels, but its ability to reveal larger structures is limited. We showed that the revised formulation avoids \emph{random} placement of points in both the synthetic and the pancreas dataset. The original ct-SNE performs better on the pancreas data than on the synthetic data, because most local neighborhoods also contain cells with a different technology label---which is not the case in the synthetic data. For cells sequenced with \tikzcircle{celseq}~celseq, there are on average 68 out of 150 nearest neighbors with a different technology label (median 63). However, \tikzcircle{fluidigmc1}~fluidigmc1 cells only have on average 10 out of 150 (median 0) such neighbors. 
This explains their \emph{almost random} embedding in Figure \ref{fig:panc_ctsne} and \ref{fig:panc_ctsne_celltypes} where the fluidigmc1, beta cells are mixed with alpha cells and fluidigmc1, alpha cells are mixed with beta cells. 

Secondly, we reflect on the two variants of \ourmethod where the desired bandwidth of the Gaussian is either estimated on $p_i$ or the final $r_i$. While the former seems justified for having a unique solution, the \emph{effective} perplexity of $r_i$ can differ from the user-defined perplexity. In the visualizations of the pancreas data (Figures S4 and S5 in the supplement) we noticed circular patterns due to a too small perplexity. We found out that this dataset contains several outlier cells that dominate the $r_i$ similarity distribution of neighboring cells with a different technology label. We did not observe these patterns when embedding the second biological dataset (Figures S1 and S2).

A final aspect is that \ourmethod redefines the similarities with the same $\beta$ for all labels. If the local HD neighborhood is already mixed with respect to the provided class labels, a larger $\beta$ might be sufficient. For cells where the distance gap between same and differently-labeled neighbors is large, a smaller $\beta$ is necessary to reweigh the similarities. In Figure \ref{fig:panc_embeddings} we show the embeddings with $\beta = \expnumber{1}{\text{-}30}$ which might overshoot the goal for some cells. That means, a too small $\beta$ can remove the neighborhood information of same-labeled points completely. In summary, neither computing the variance on $p_i$ nor on $r_i$ ensures a stable balance between same and differently-labeled neighbors.

\vspace{4mm}\pstart{Future Work}
Based on the result that \ourmethod addresses some shortcomings of ct-SNE but brings a different set of drawbacks, we see various avenues for future work. Firstly we assume that the \rnx scores for \ourmethod embeddings would increase when the same-labeled similarities do not vanish for small $\beta$. Additionally, one could allow for separate \textit{merging strengths} for every label, based on the assumption that all labels should be mixed in the resulting embedding. 
To ensure a certain trade-off between effective same and differently-labeled neighbors one could compute two separate similarity distributions with two different perplexities. This is a similar idea as implemented in Class-aware t-SNE \cite{Bodt2019ClassawareTC} where the perplexity is adjusted to reach a certain ratio of same-labeled points in the neighborhood.
Finally, the embeddings by ct-SNE and \ourmethod could be compared to the reference embeddings by \citet{Policar2021} or a combination of data integration methods for biological datasets and t-SNE.

\section{Conclusion}
We presented \ourmethod to find low-dimensional embeddings that show structure beyond a previously known clustering. Conditional t-SNE can fail to reveal this structure when focusing only on the local neighborhoods using small perplexities. To resolve this limitation, we reformulate the original idea to condition the high-dimensional similarities and explicitly include nearest neighbors with different labels. Our experiments on synthetic data confirmed that the \ourmethod improved on the criticized aspects of the original method, but ct-SNE performed better in terms of label mixing and neighborhood preservation on real-world single-cell data. Finally, we investigated limitations of \ourmethod and proposed to control the number of effective same and differently-labeled neighbors more explicitly.

\subsubsection{Acknowledgements} The authors would like to thank Tijl De Bie and Yvan Saeys for the discussions about conditional t-SNE. This research was funded by the ERC under the EU's 7th Framework and H2020 Programmes (ERC Grant Agreement no. 615517 and 963924), the Flemish Government (AI Research Program), and the FWO (project no. 11J2322N, G0F9816N, 3G042220).

\pagebreak
\setcounter{equation}{0}
\setcounter{figure}{0}
\setcounter{table}{0}
\setcounter{page}{1}
\makeatletter
\renewcommand{\theequation}{S\arabic{equation}}
\renewcommand{\thefigure}{S\arabic{figure}}
\title{Supplement \\Revised Conditional t-SNE:\\Looking Beyond the Nearest Neighbors}
\author{Edith Heiter\inst{1} \and Bo Kang\inst{1} \and Ruth Seurinck\inst{12} \and Jefrey Lijffijt\inst{1}}%

\authorrunning{E. Heiter et al.}
\titlerunning{Supplement, Revised Conditional t-SNE}
\institute{Ghent University, Belgium\\
	\email{\{edith.heiter,bo.kang,ruth.seurinck,jefrey.lijffijt\}@ugent.be} \and
	VIB Center for Inflammation Research, Belgium}
\maketitle              %
\section{Human Immune Dataset}
The \href{https://figshare.com/ndownloader/files/25717328}{human immune dataset}, aggregated by \citet{luecken2022benchmarking} contains single cell gene expression measurements of  $n=33506$ cells along $d=12303$ genes. It is a combination of ten batches with cells from five donors, four sequencing technologies and two tissues (peripheral blood and bone marrow).
It also includes annotations for 16 different celltypes. We perform the same preprocessing and use their \href{https://scib.readthedocs.io/en/latest/index.html}{scib}\cite{luecken2022benchmarking} package for batch-aware highly-variable gene selection and scaling. Finally, we project the 2000 highly-variable genes to 50 dimensions with PCA.

With ct-SNE \cite{kang2021conditional} and \ourmethod, we aim to factor out the differences related to the \textit{batch} labels. For all embeddings we use a perplexity of $50$, $\theta = 0.5$, and $1000$ iterations. This took about 470 seconds with ct-SNE and 80 seconds with \ourmethod (2xIntel(R) Xeon(R) Gold 6136 CPU @ 3.00GHz). The quality was assessed on a subset of 5\% of the data (1675 cells) using the adjusted RNX and Laplacian scores.

\begin{figure}
	\begin{subfigure}{\textwidth}
		\includegraphics[width=\textwidth]{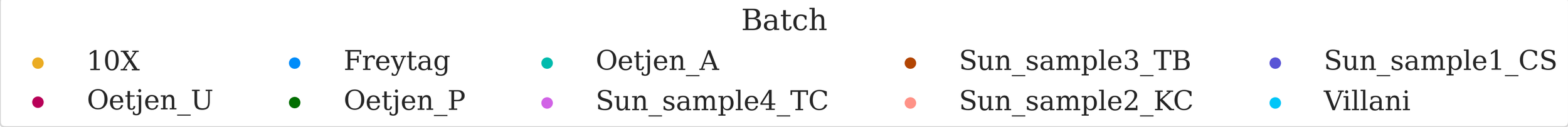}
	\end{subfigure}
	\begin{subfigure}[b]{.32\textwidth}
		\includegraphics[width=\textwidth]{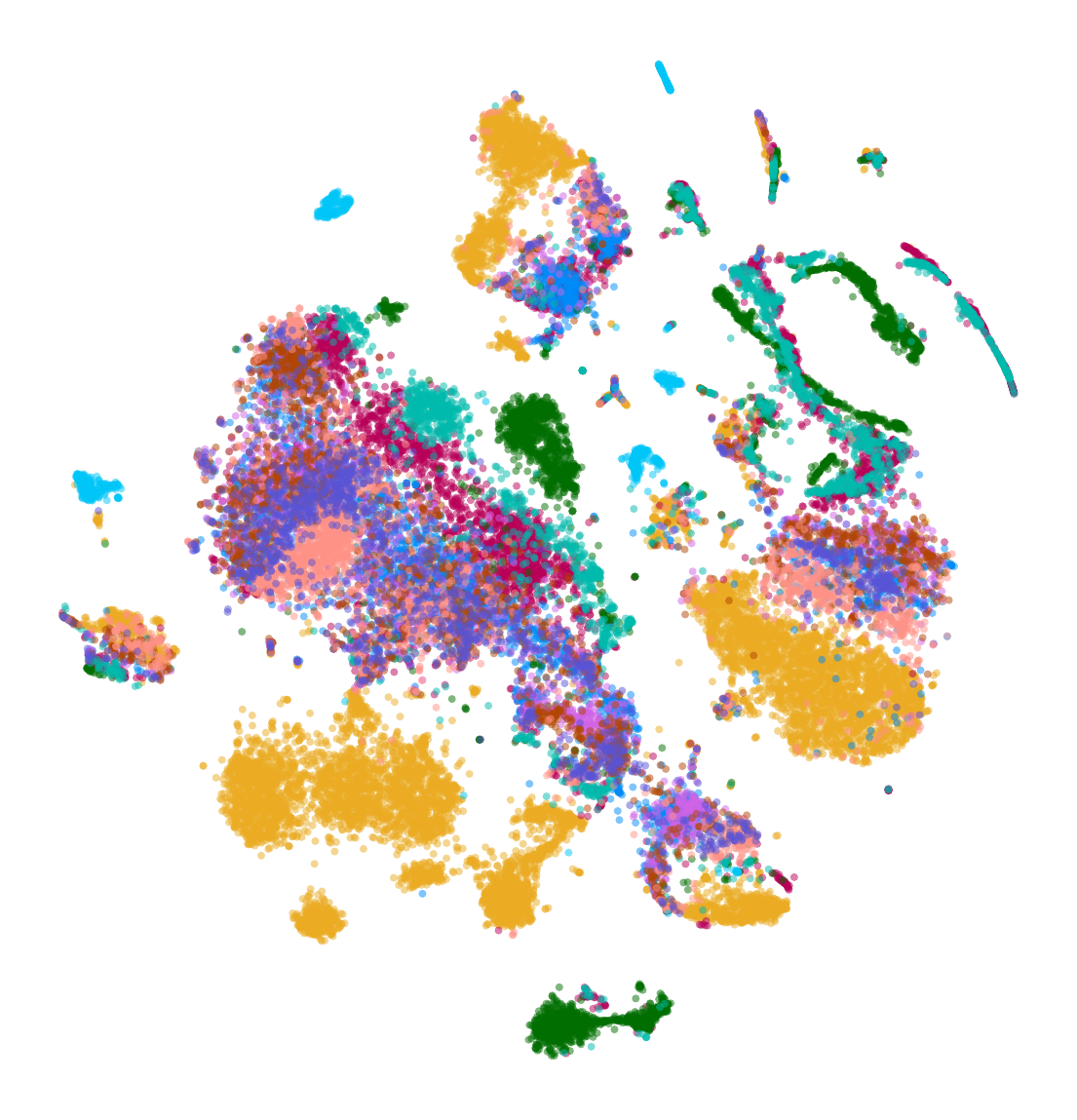}
		\caption{t-SNE}
		\label{fig:immune_tsne}
	\end{subfigure}\hfill
	\begin{subfigure}[b]{.32\textwidth}
		\includegraphics[width=\textwidth]{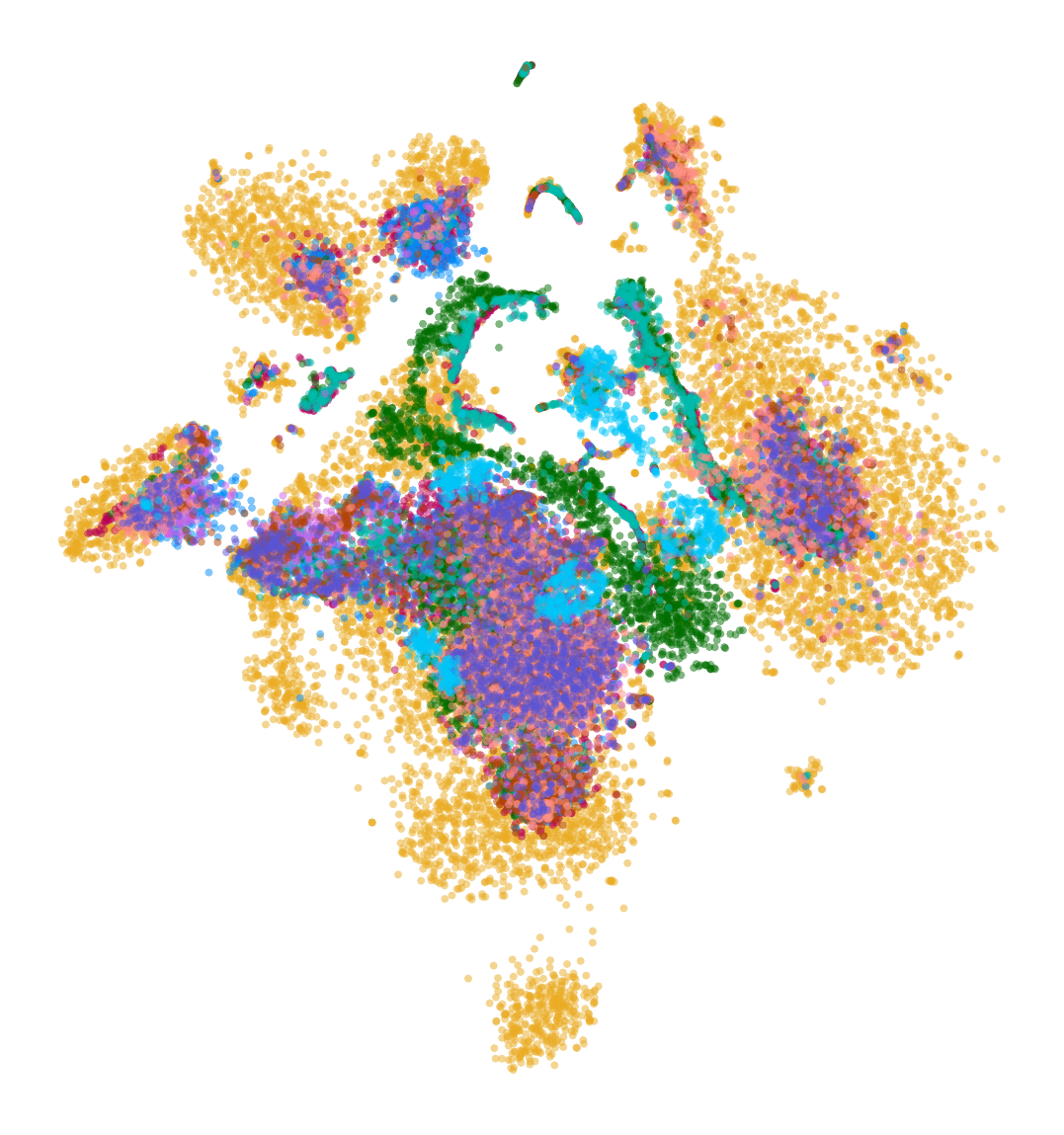}
		\caption{ct-SNE}
		\label{fig:immune_ctsne}
	\end{subfigure}\hfill
	\begin{subfigure}[b]{.32\textwidth}
		\includegraphics[width=\textwidth]{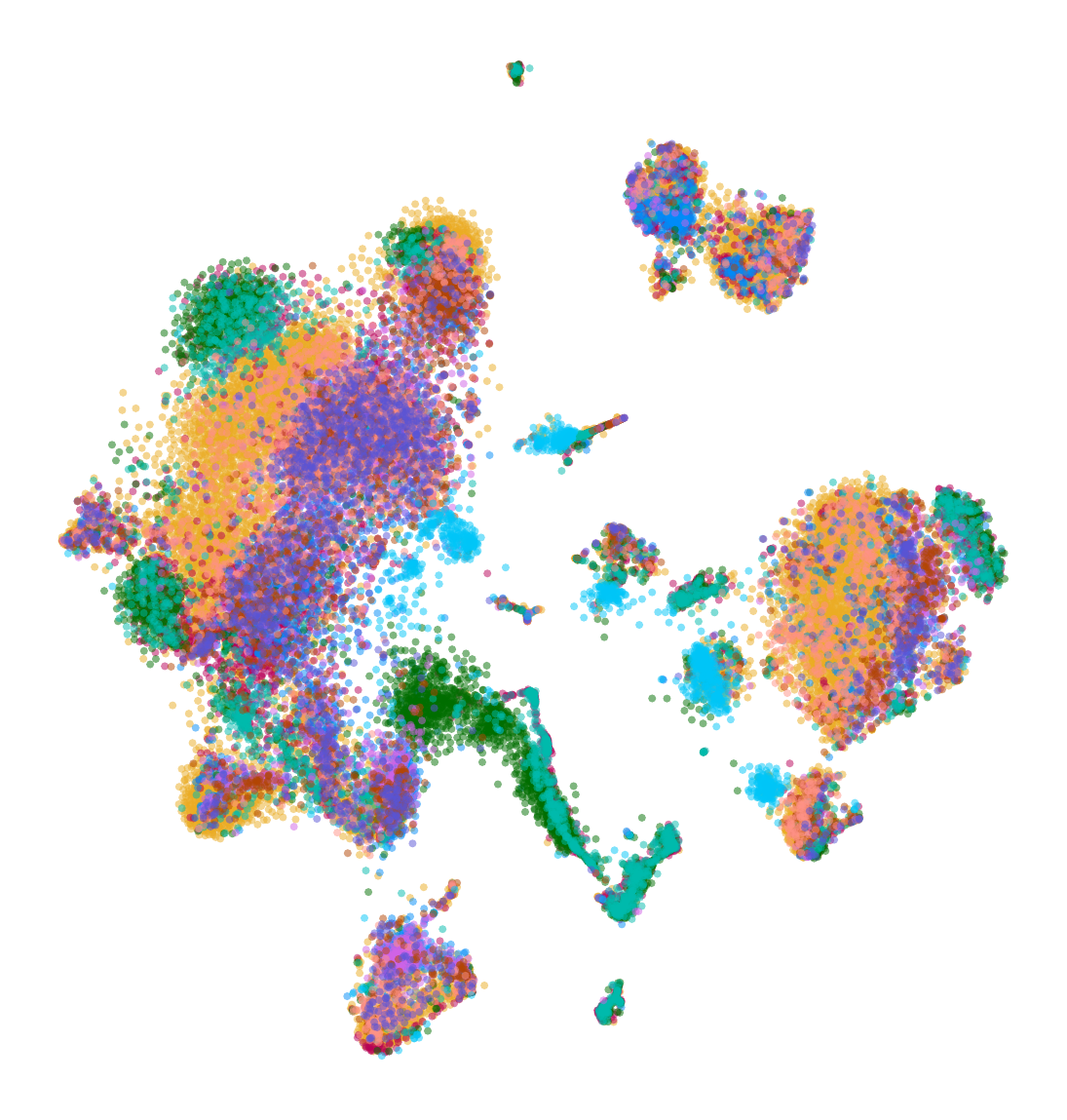}
		\caption{\ourmethod (var. $r_i$)}
		\label{fig:immune_fastctsne}
	\end{subfigure}\hfill
	\begin{subfigure}[b]{.32\textwidth}
		\includegraphics[width=\textwidth]{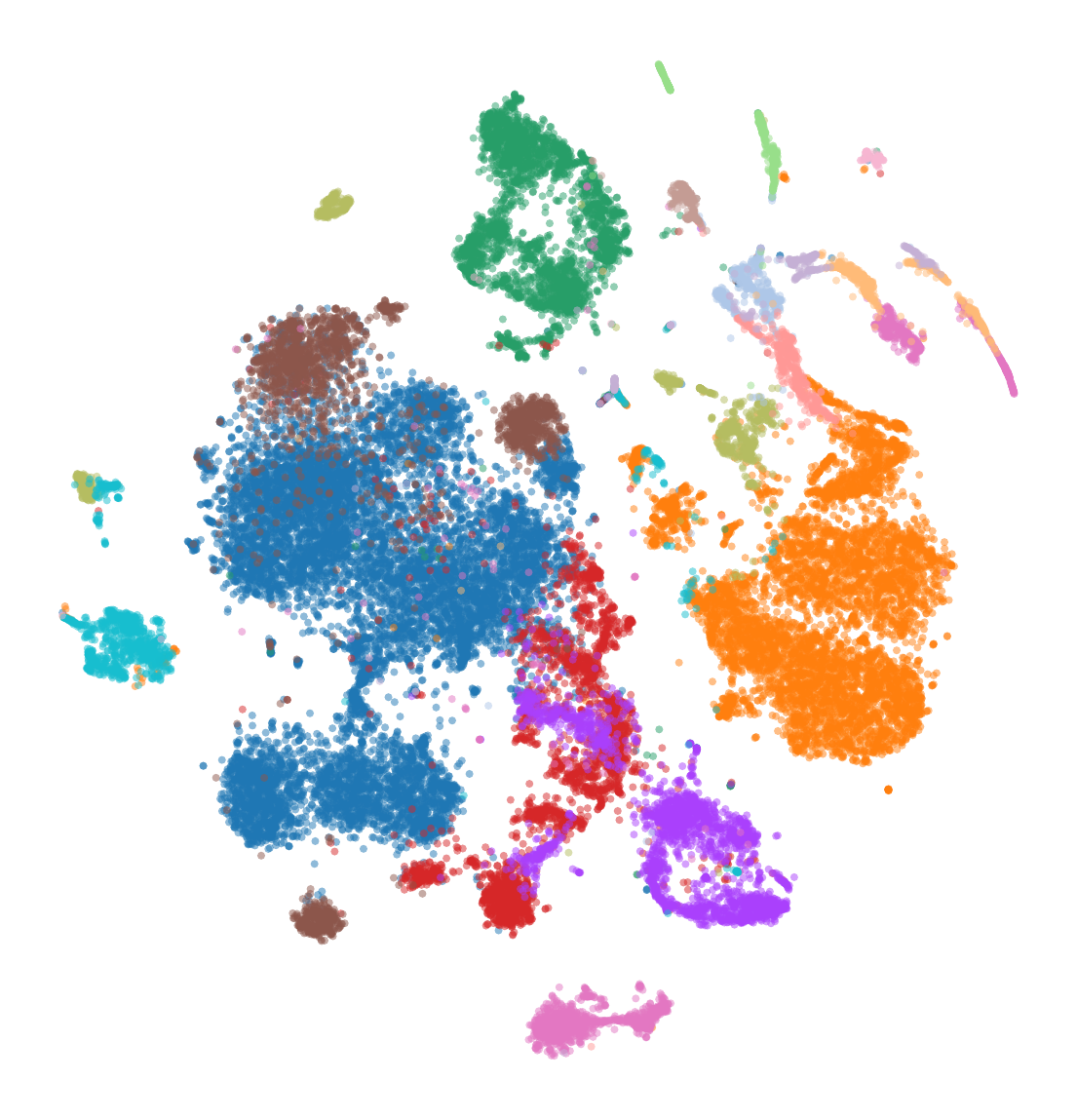}
		\caption{t-SNE}
		\label{fig:immune_tsne_celltypes}
	\end{subfigure}\hfill
	\begin{subfigure}[b]{.32\textwidth}
		\includegraphics[width=\textwidth]{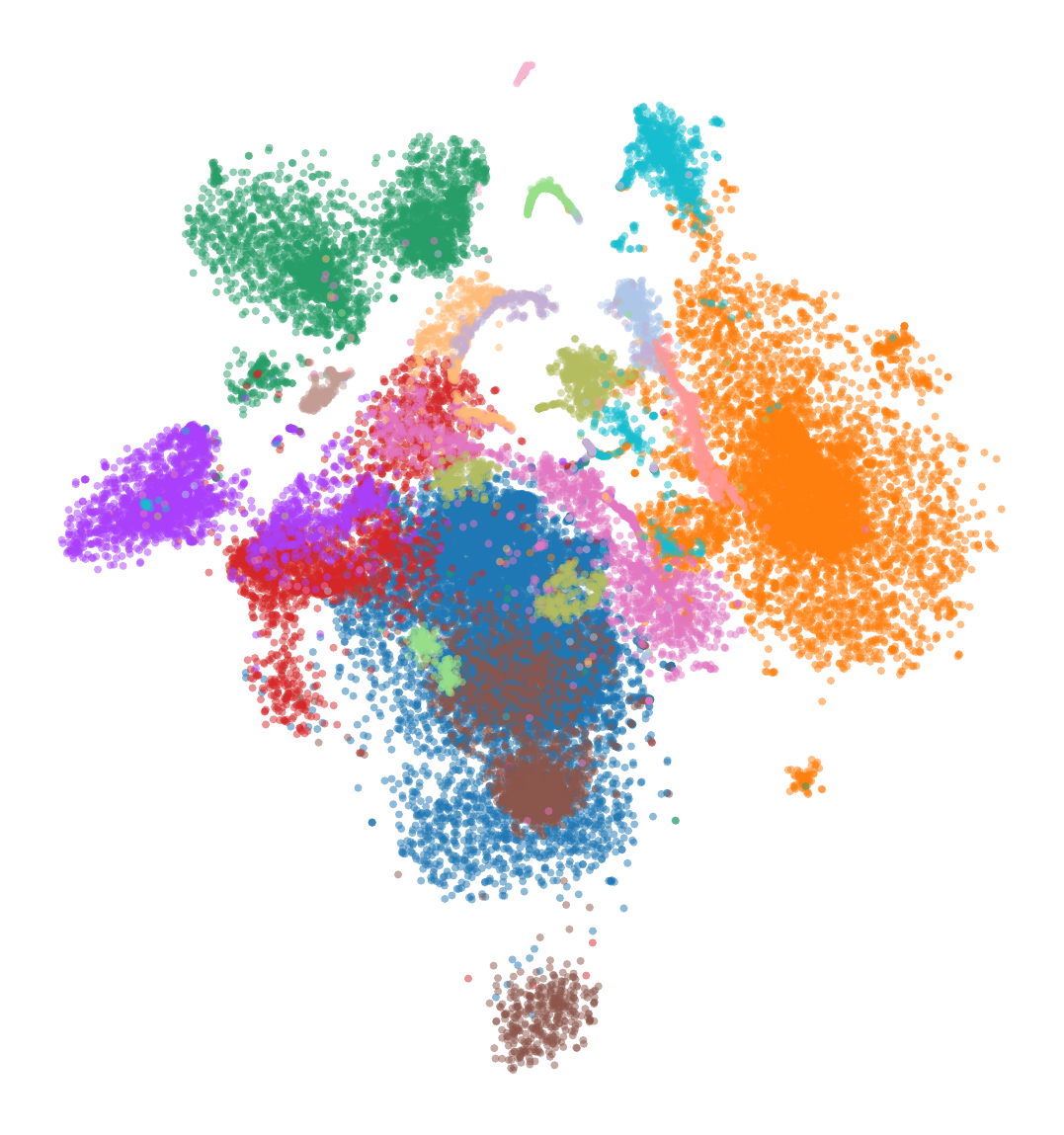}
		\caption{ct-SNE}
		\label{fig:immune_ctsne_celltypes}
	\end{subfigure}\hfill
	\begin{subfigure}[b]{.32\textwidth}
		\includegraphics[width=\textwidth]{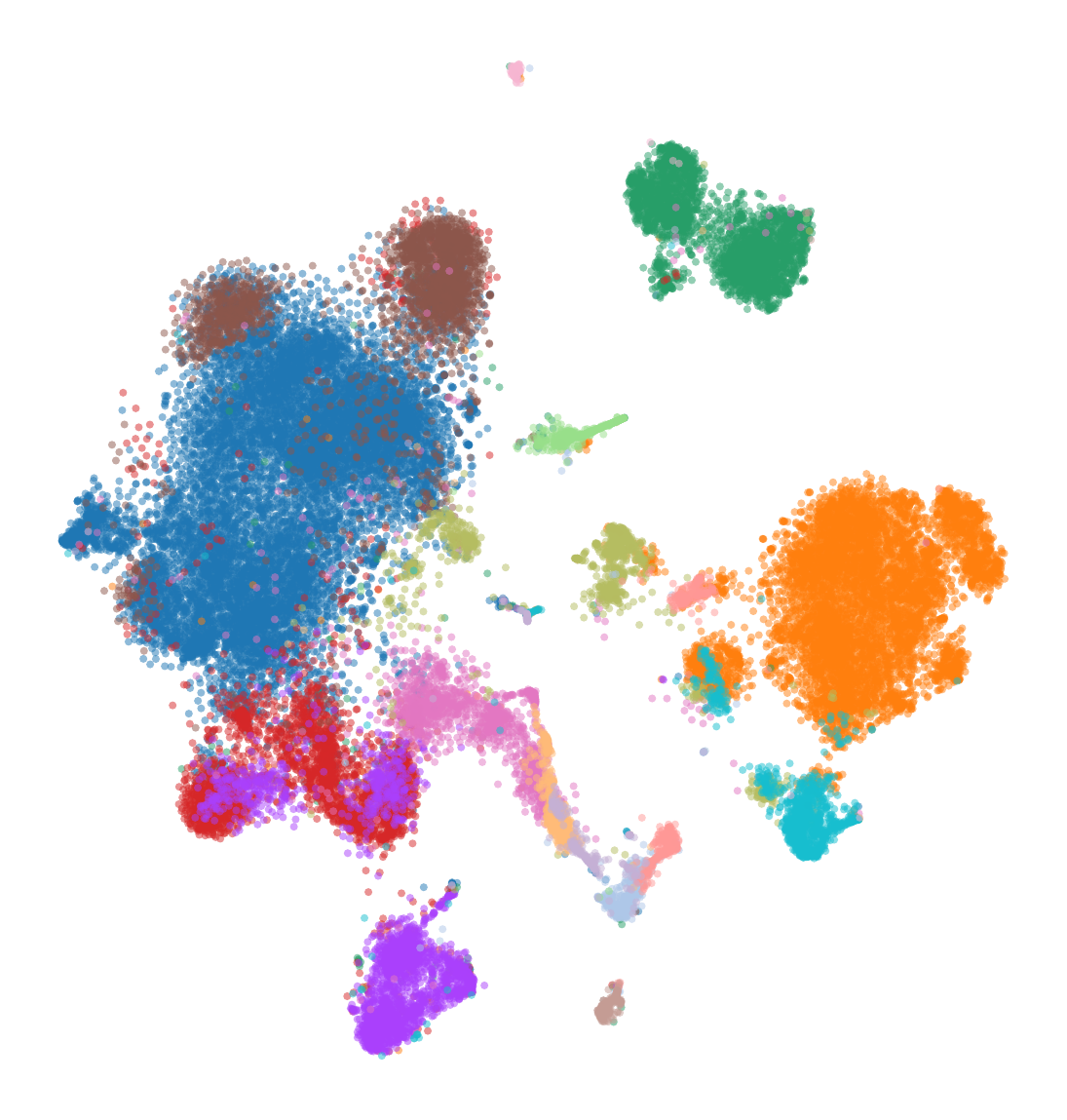}
		\caption{\ourmethod (var. $r_i$)}
		\label{fig:immune_fastctsne_celltypes}
	\end{subfigure}
	\begin{subfigure}{\textwidth}
		\vspace{.5cm}
		\includegraphics[width=\textwidth]{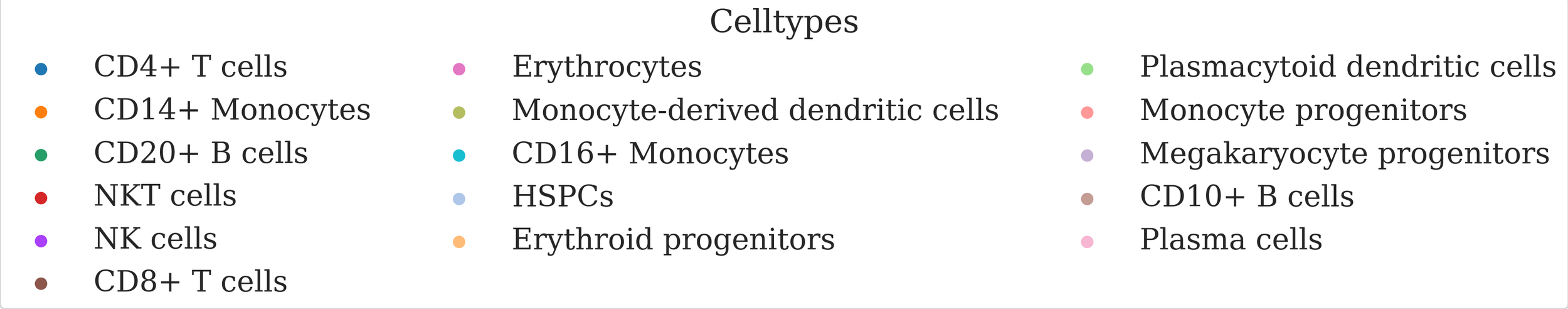}
	\end{subfigure}
	\caption{Visualizations of the Human Immune data where the batch labels are provided as prior information to ct-SNE and \ourmethod. For ct-SNE and \ourmethod we visualize the embeddings with $\beta = \expnumber{1}{\text{-}8}$. Cell coloring according to batch (\ref{fig:immune_tsne})-(\ref{fig:immune_fastctsne}) and celltype (\ref{fig:immune_tsne_celltypes})-(\ref{fig:immune_fastctsne_celltypes}).}
	\label{fig:immune_embeddings}
\end{figure}

\begin{figure}
	\begin{subfigure}{.45\textwidth}
		\includegraphics[width=\textwidth]{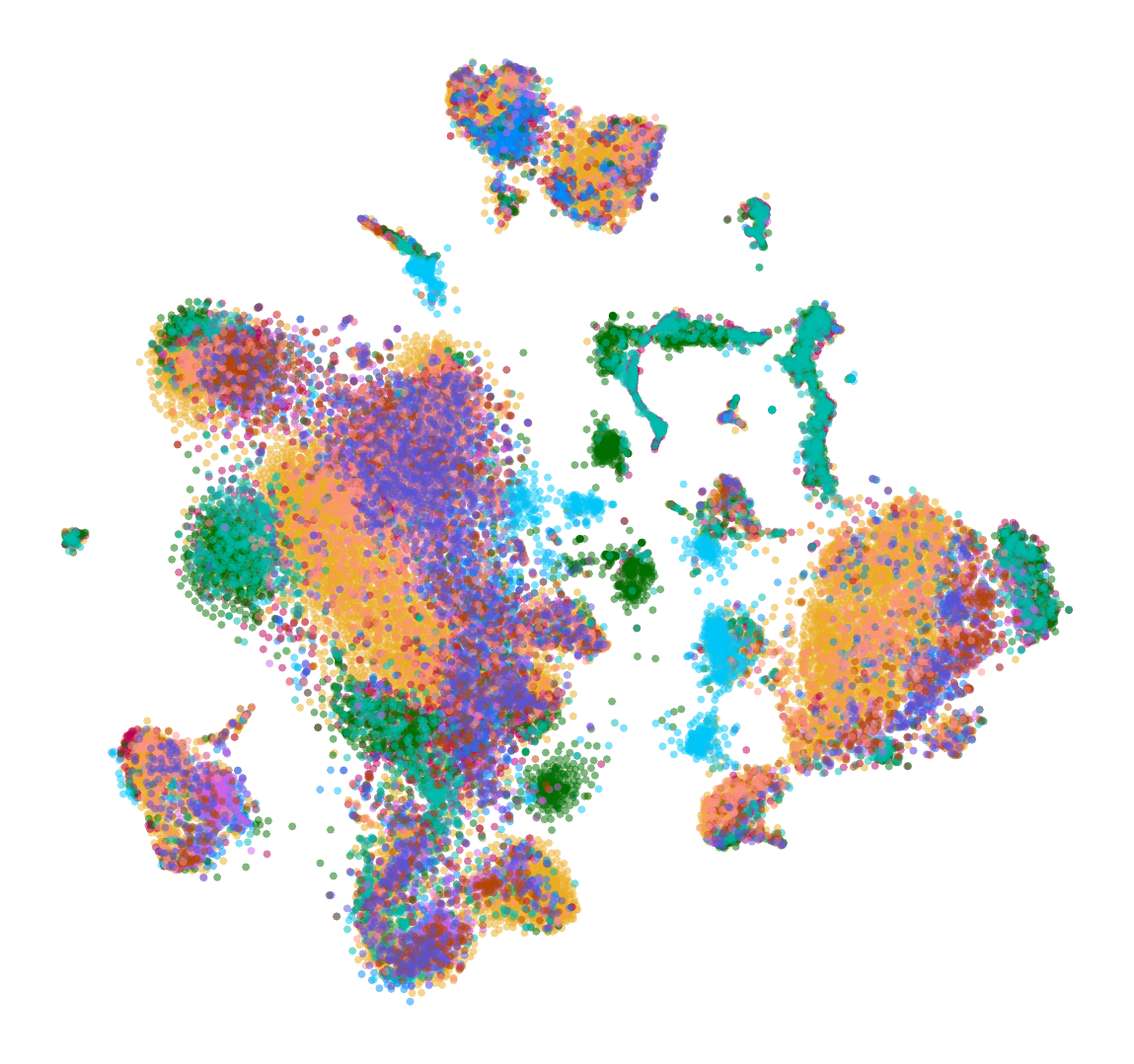}
		\caption{\ourmethod (var. $p_i$), coloring by batch}
	\end{subfigure}\hfill
	\begin{subfigure}{.45\textwidth}
		\includegraphics[width=\textwidth]{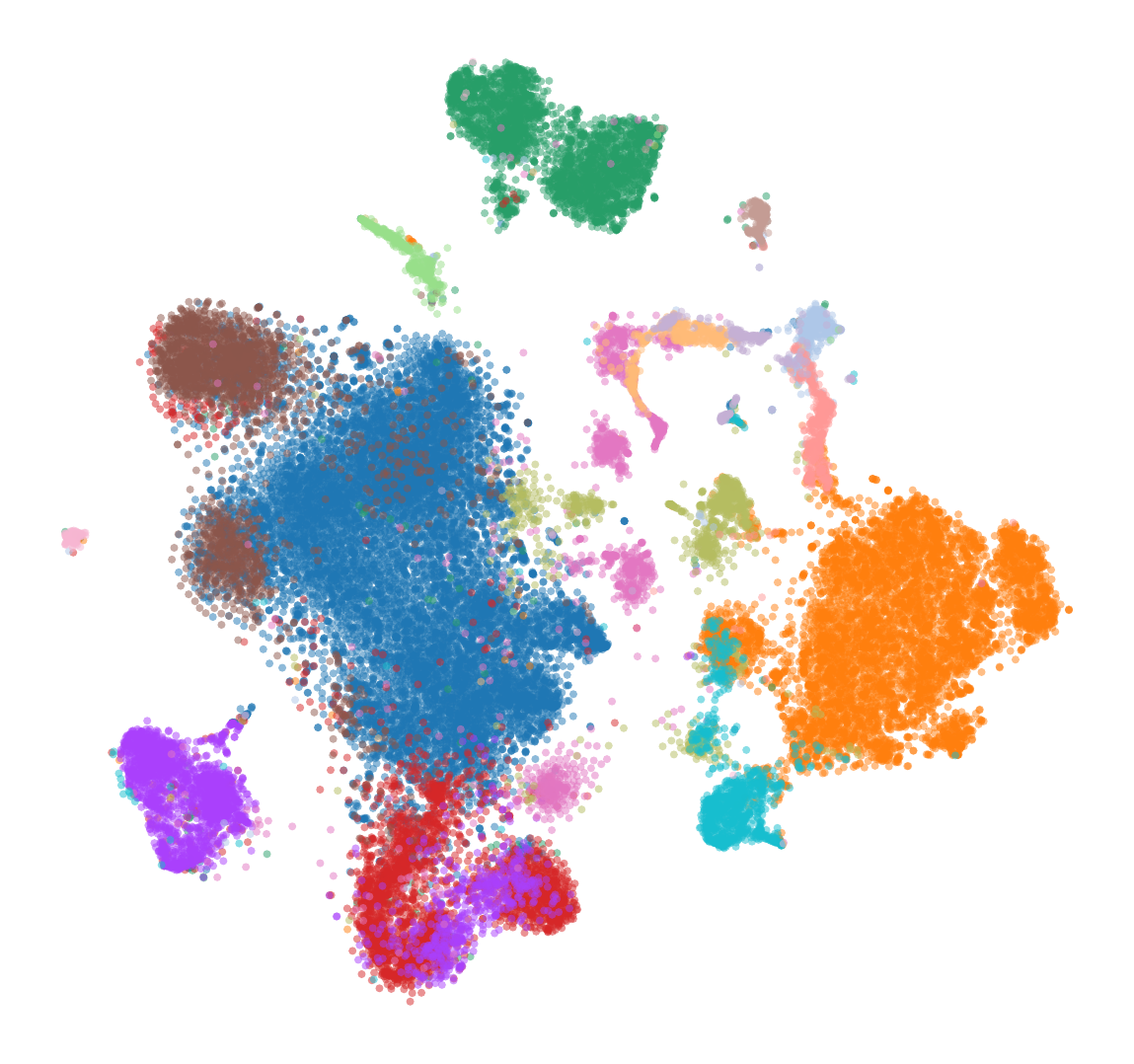}
		\caption{\ourmethod (var. $p_i$), coloring by celltype}
	\end{subfigure}
	\caption{Embeddings of the immune dataset by \ourmethod with variance estimation on $p_i$}
	\label{fig:immune_fastctsne_embeddings}
\end{figure}

\begin{figure}
	\begin{subfigure}{\textwidth}
		\includegraphics[width=\textwidth]{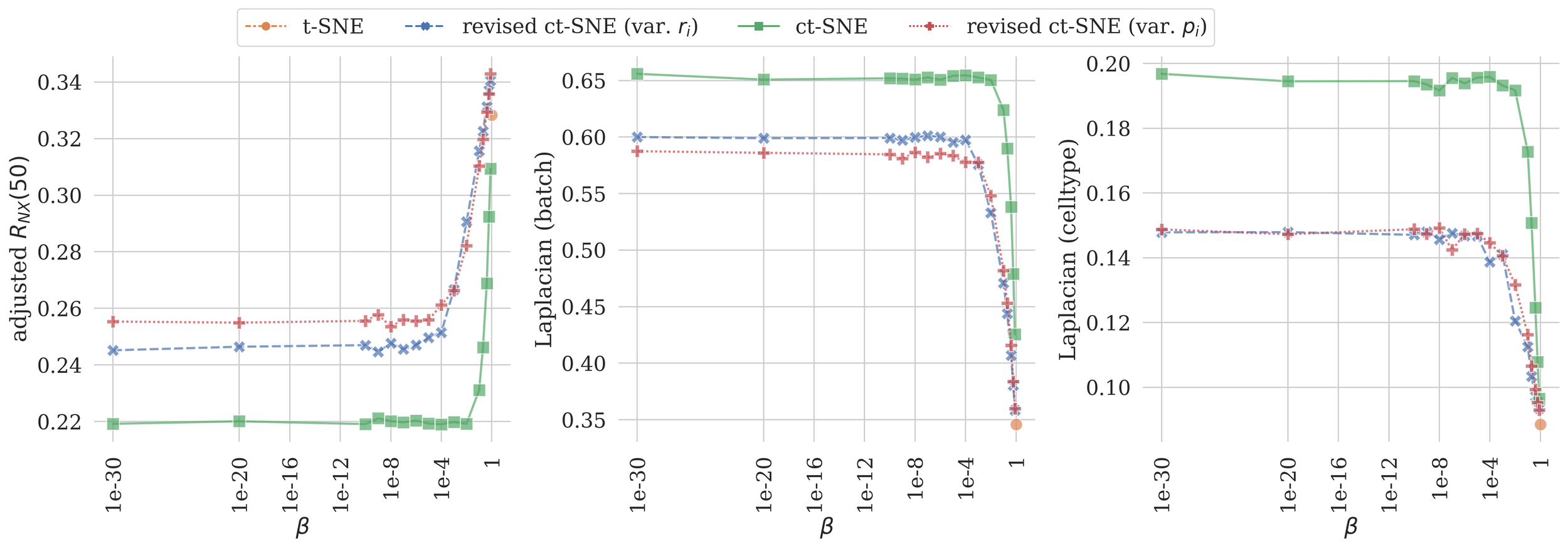}
	\end{subfigure}
	\begin{subfigure}{\textwidth}
		\hfill
		\includegraphics[width=.33\textwidth]{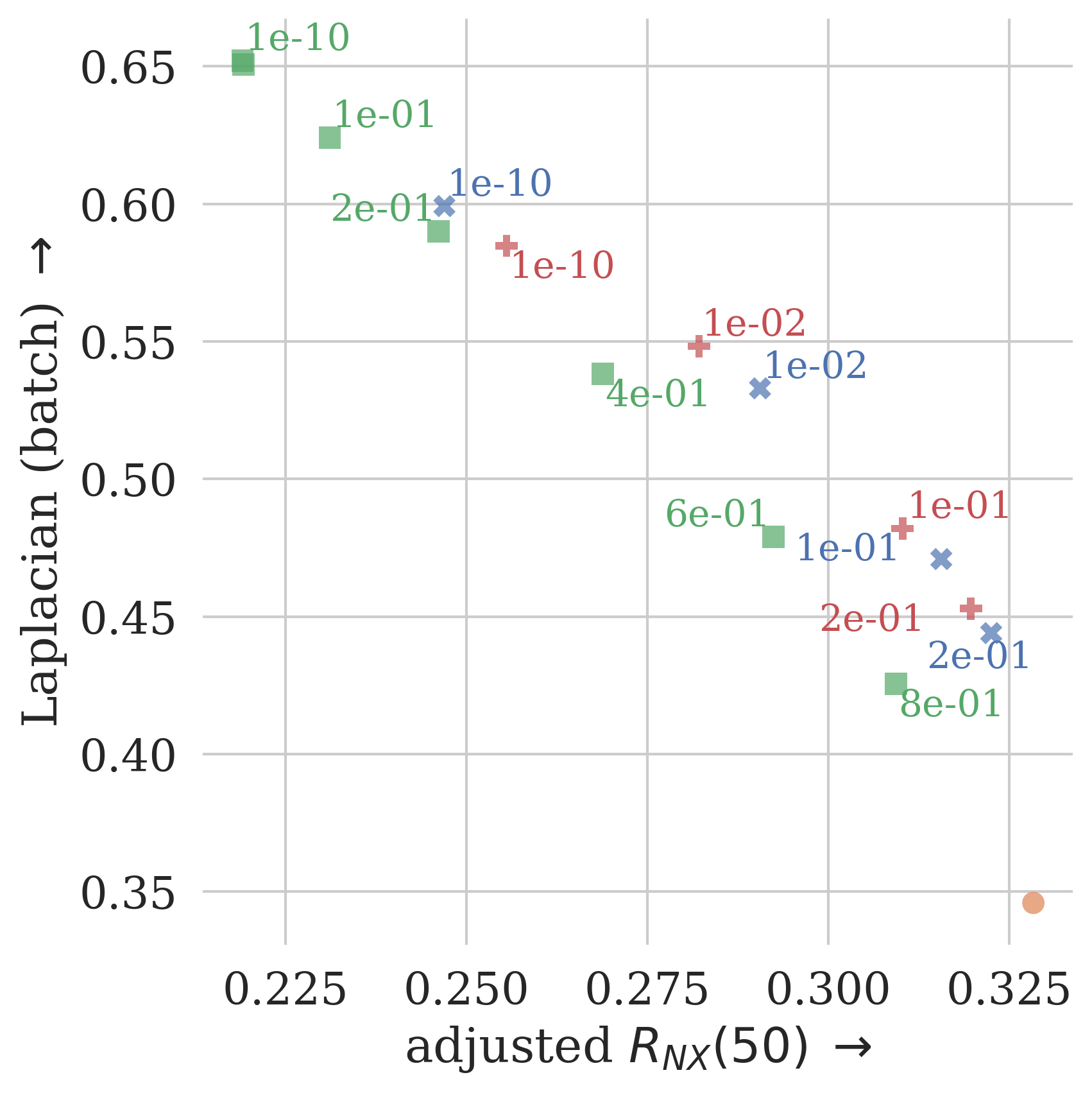}
		\includegraphics[width=.33\textwidth]{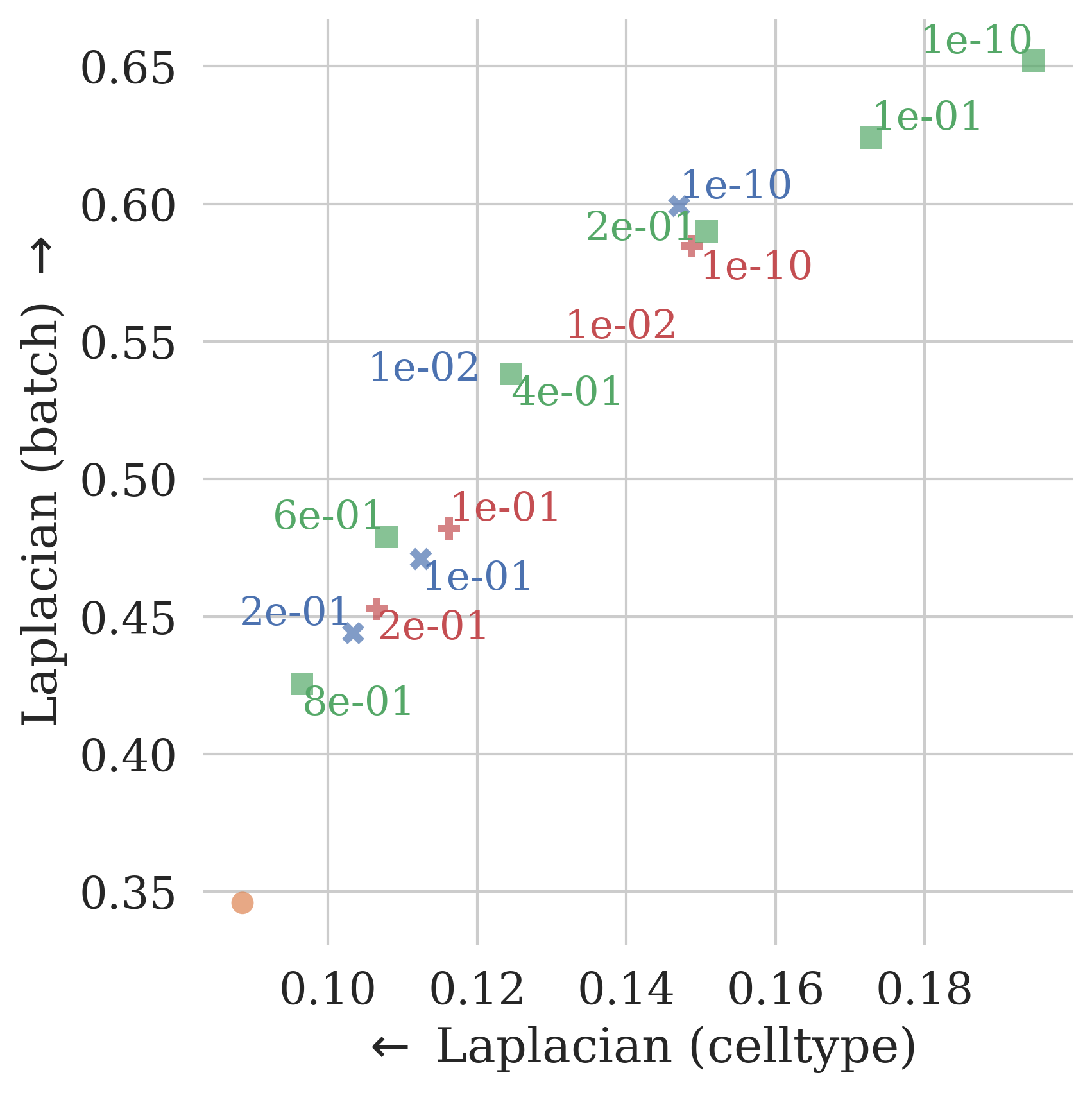}
	\end{subfigure}
	\caption{Human Immune dataset embeddings evaluated with a neighborhood size of $k=50$. The t-SNE scores are depicted in orange at $\beta = 1$. The first two rows show the trend of the evaluation measures when changing the value for $\beta$. The bottom row shows the trade-offs between different evaluation criteria.}
	\label{fig:immune_evaluation}
\end{figure}

\begin{table}
	\centering
	\caption{The cells of the datasets were processed using one of three sequencing technologies and stem either from bone marrow or blood samples (PBMC).}
	\label{tab:immune_labels}
	\begin{tabular}{lll}
		batch & technology & tissue \\
		\toprule
		10X & v3\_10X & PBMC \\
		Freytag & v2\_10X & PBMC \\
		OetjenA & v2\_10X & bone marrow \\
		OetjenP & v2\_10X & bone marrow \\
		OetjenU & v2\_10X & bone marrow \\
		Sun1 & 10X & PBMC \\
		Sun2 & 10X & PBMC \\
		Sun3 & 10X & PBMC \\
		Sun4 & 10X & PBMC \\
		Villani & smart-seq2 & PBMC
	\end{tabular}
\end{table}

In Figures \ref{fig:immune_embeddings} and \ref{fig:immune_fastctsne_embeddings} we show the embeddings of t-SNE, ct-SNE, and \ourmethod coloring the cells according to their batch label in the top row and celltype in the bottom row. The visualization by t-SNE clearly shows batch effects. Cells from \textit{Villani} and \textit{10X} are separate from all other batches. They were also processed using a different technology (see Table \ref{tab:immune_labels}). Also cells from \textit{Sun} en \textit{Freytag} cluster together in t-SNE and were sequenced using v2\_10X and 10X. The cells from \textit{Oetjen} are also slightly separate from the other batches and are the only ones that stem from bone marrow. 

Comparing the embeddings of ct-SNE and \ourmethod in Figure \ref{fig:immune_embeddings}, we observe that both versions reduce the separation of batches. \Ourmethod results in more compact clusters and the coloring according to celltype reveals that ct-SNE results in overlapping celltype clusters.
From the evaluation measures in Figure \ref{fig:immune_evaluation} we conclude that ct-SNE is able to merge cells from different batches but does so at the expense of keeping the celltype clusters separate. \Ourmethod slightly increases the number of cells with different celltype label in the neighborhood and leads to a better neighborhood preservation for similar Laplacian (batch) values.

\section{Pancreas Dataset}

To support the analysis of ct-SNE and \ourmethod, we analyze the high-dimensional neighborhoods. As ct-SNE is based on the similarity information based on a local neighborhood, we visualize the fraction of neighbors with different technology label in Figures \ref{fig:panc_neighbors_celltype} and \ref{fig:panc_neighbors_technology}. We note that the indrop cells have very few differently-labeled neighbors which could be the reason of them not being merged with cells from other technologies in the ct-SNE embedding in Figure 6c of the main paper.

\begin{figure}
	\includegraphics[width=\textwidth]{pancreas/pancreas_tech_legend.png}
	\begin{subfigure}[b]{.45\textwidth}
		\includegraphics[width=\textwidth]{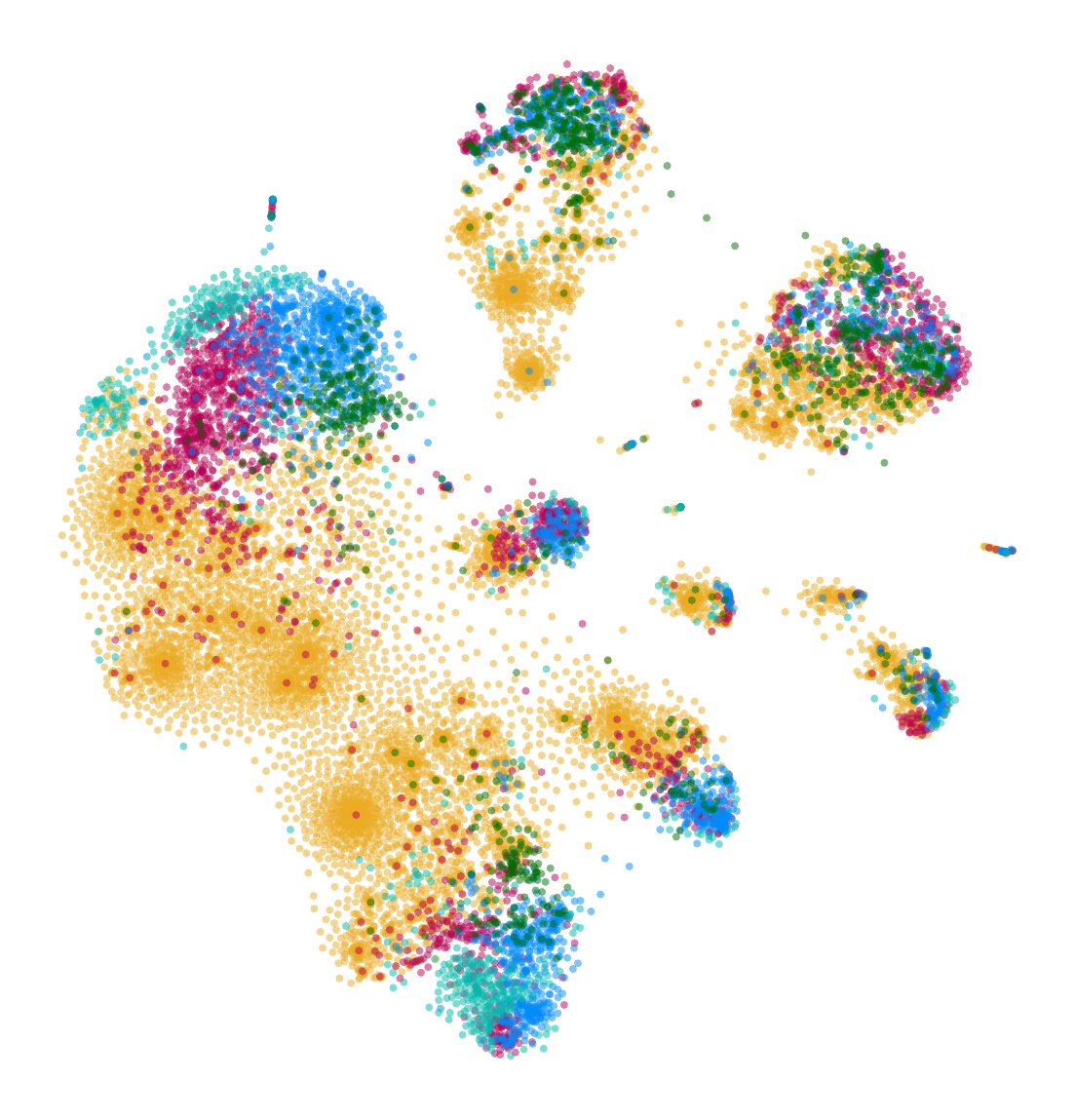}
		\caption{\Ourmethod (var. $p_i$)}
		\label{fig:panc_fastctsne_varfirst_tech}
	\end{subfigure}\hfill
	\begin{subfigure}[b]{.45\textwidth}
		\includegraphics[width=\textwidth]{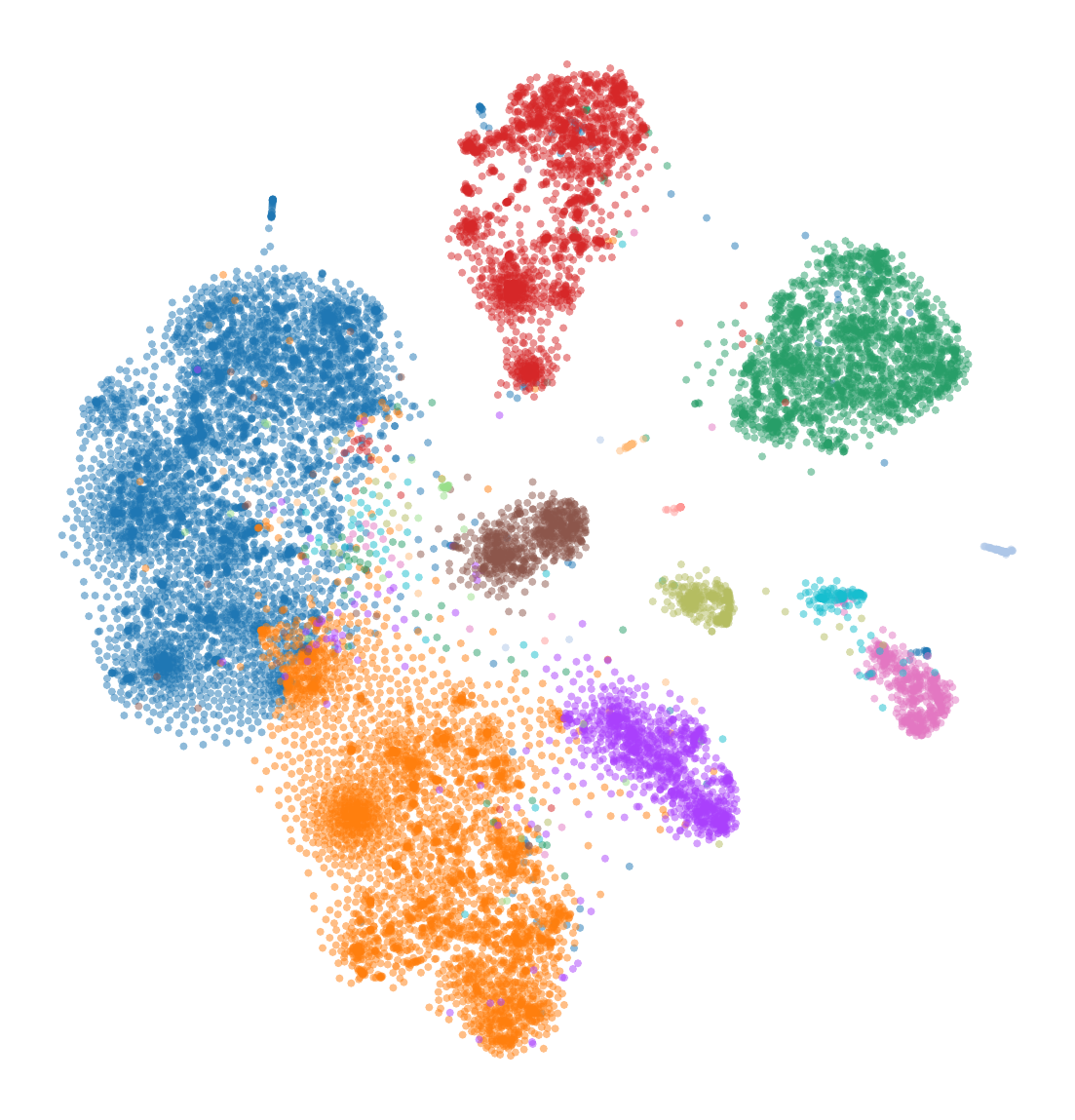}
		\caption{\Ourmethod (var. $p_i$)}
		\label{fig:panc_fastctsne_varfirst_celltypes}
	\end{subfigure}
	\includegraphics[width=\textwidth]{pancreas/pancreas_celltype_legend.png}
	\caption{Embeddings of the pancreas data by \ourmethod with variance estimation on $p_i$.}
	\label{fig:panc_fastctsne_varfirst}
\end{figure}

\begin{figure}
	\begin{subfigure}{.6\textwidth}
		\includegraphics[width=\textwidth]{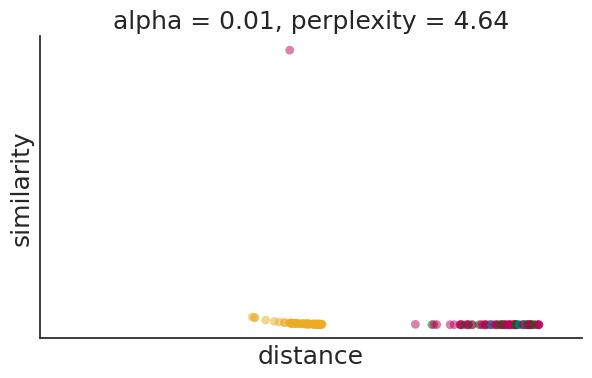}
		\caption{HD similarities $r_i$ for a \tikzcircle{indrop} indrop cell when fixing the variance on $p_i$ first. The final similarity distribution $r_i$ has a small perplexity due to the \tikzcircle{smartseq2} smartseq2 outlier point in the neighborhood.}
	\end{subfigure} \hfill
	\begin{subfigure}{.38\textwidth}
		\includegraphics[width=\textwidth]{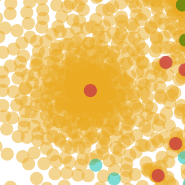}
		\caption{Embedding structure where many \tikzcircle{indrop} indrop cells have a high pairwise $r_{ij}$ similarity towards the same \tikzcircle{smartseq2} smartseq2 cell.}
		\label{fig:panc_pattern}
	\end{subfigure}
	\caption{Similarity distribution with low perplexity due to outliers in (a) and final embedding of \ourmethod (cropped from Figure 4a) in (b).}
	\label{fig:outlier_embedding}
\end{figure}

\begin{figure}
	\includegraphics[width=\textwidth]{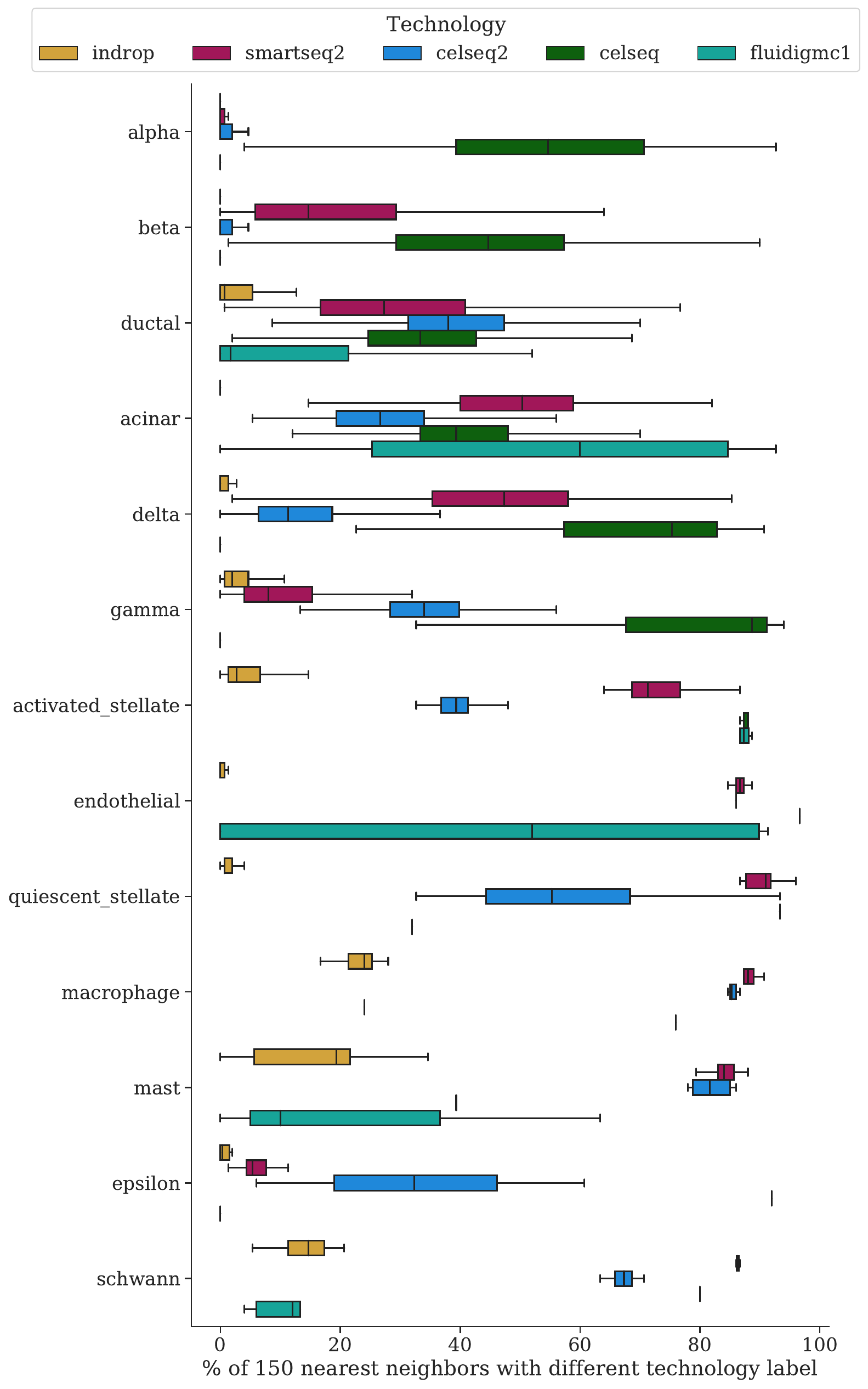}
	\caption{Boxplots showing the percentage of nearest neighbors in the pancreas dataset with different technology label grouped per celltype and colored by technology. The length of the whiskers is limited to $1.5 \cdot \text{interquartile range}$ and we do not show outliers.}
	\label{fig:panc_neighbors_celltype}
\end{figure}

\begin{figure}
	\includegraphics[width=.9\textwidth]{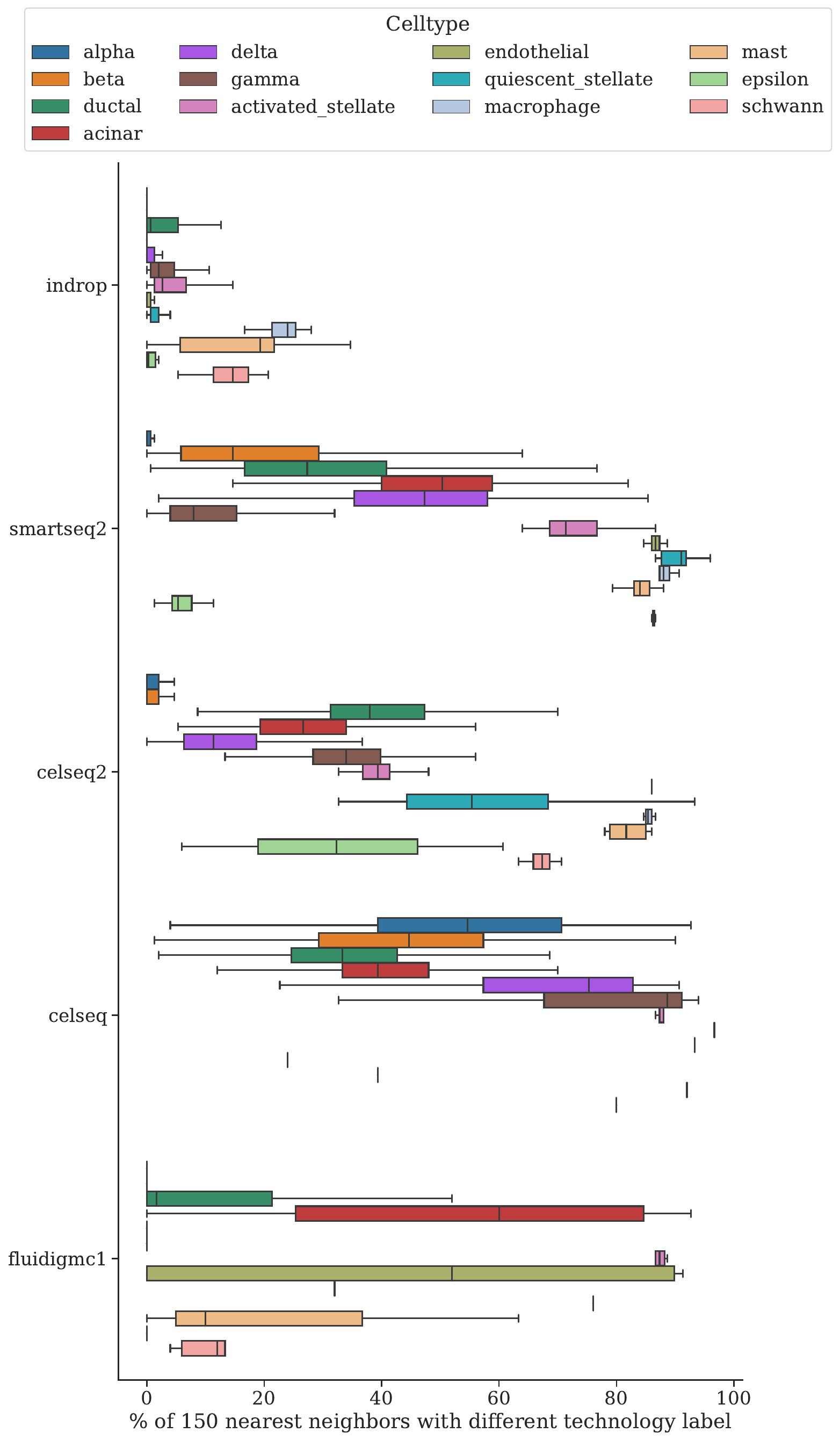}
	\caption{Boxplots showing the percentage of nearest neighbors in the pancreas dataset with different technology label grouped by technology and colored by celltype. The length of the whiskers is limited to $1.5 \cdot \text{interquartile range}$ and we do not show outliers.}
	\label{fig:panc_neighbors_technology}
\end{figure}

We show the \ourmethod embedding where the variance is estimated on $p_i$ in Figure \ref{fig:panc_fastctsne_varfirst} and observe several regions with the same pattern as in Figure \ref{fig:panc_pattern}:  
For some \tikzcircle{indrop} indrop cells i, the $r_i$ similarities are dominated by the same \tikzcircle{smartseq2} smartseq2 cell j. This smartseq2 outlier has a high similarity $p_{j_{\tikzcircle{smartseq2}}| i_{\tikzcircle{indrop}}}$ that only gets larger when the similarity to all differently-labeled points is multiplied with $\alpha > 1$. After re-normalization, all other $r_{\cdot|i}$ will be almost zero and the effective perplexity of $r_i$ is very small.

\bibstyle{splncs04}
\bibliography{references.bib}

\end{document}